\documentclass{article} 
\usepackage{iclr2027_conference,times}

\usepackage{amsmath,amsfonts,bm}

\def\eqref#1{equation~\ref{#1}}

\def\1{\bm{1}}

\DeclareMathAlphabet{\mathsfit}{\encodingdefault}{\sfdefault}{m}{sl}
\SetMathAlphabet{\mathsfit}{bold}{\encodingdefault}{\sfdefault}{bx}{n}

\usepackage{hyperref}
\usepackage{url}

\usepackage{graphicx}
\usepackage{amsmath}
\usepackage{amssymb}
\usepackage{booktabs}
\usepackage{capt-of} 
\usepackage{pifont}
\usepackage{multirow}
\usepackage[table]{xcolor}

\usepackage{algorithm}
\usepackage{algorithmic}
\usepackage{amsmath}

\newcommand{\cmark}{\ding{51}}
\newcommand{\xmark}{\ding{55}}

\def\eg{\emph{e.g.,~}}
\def\ie{\emph{i.e.,~}}
\def\ournet{RoXDrive}

\hypersetup{
     colorlinks=true,
     linkcolor=red,
     citecolor=cyan,
     filecolor=magenta, 
     urlcolor=cyan,
     }

\title{RoXDrive: Closed-Loop Reinforcement Learning for End-to-End Autonomous Driving via Action-Faithful Rollouts}

\author{\textbf{Hongbin Lin}$^{1}$\thanks{Authors contributed equally. },
\textbf{Chaoda Zheng}$^{2}$\footnotemark[1],
\textbf{Yiming Yang}$^{1}$,
\textbf{Xiangyu Li}$^{2}$,
\textbf{Shijia Chen}$^{2}$,
\textbf{Jinhao Deng}$^{2}$, \\
\textbf{Kangjie Chen}$^{2}$,
\textbf{Dongbin Zhang}$^{2}$,
\textbf{Jie Feng}$^{3}$,
\textbf{Yu Zhang}$^{2}$,
\textbf{Xianming Liu}$^{2}$,
\textbf{Shuguang Cui}$^{1}$, \\
\textbf{Boyang Wang}$^{2}$\thanks{Corresponding authors.},
\textbf{Zhen Li}$^{1}$\footnotemark[2]\\[0.4em]
$^{1}$The Chinese University of Hong Kong, Shenzhen\\
$^{2}$XPeng Motors\\
$^{3}$Xidian University
}

\iclrfinalcopy 
\begin{document}

\maketitle
\lhead{}

\begin{abstract}
End-to-end autonomous driving policies are commonly trained via imitation learning on logged demonstrations without observing the consequences of their own actions, leading to causal confusion in closed-loop real-world deployment.
To address this issue, reinforcement learning (RL) post-training offers a promising alternative by leveraging world models as interactive training environments to enable future scene generation for policy improvement.
Nevertheless, existing approaches either rely on reconstruction-based simulators, offering limited counterfactual interaction, or adopt synthetic simulators to enable long-horizon closed-loop interaction at the cost of a substantial sim-to-real gap. 
Recently, video world models have exhibited the ability to generate realistic multi-step future rollouts but may not faithfully reflect action conditions, resulting in action-vision mismatch.
In this paper, we introduce \emph{\textbf{\ournet}}, a plug-and-play closed-loop RL framework that enables reliable policy optimization by identifying action-faithful world-model rollouts, consisting of two stages: 
1) Model pre-training: In addition to imitation-based policy pre-training, we devise an Action-Vision Faithfulness Evaluator for inverse dynamics estimation with our geometry-aware auxiliary trajectory supervision, enabling assessment of whether visual dynamics faithfully reflect conditioning ego actions.
2) Action-faithful RL post-training: Agents iteratively interact with world models to form long-horizon scene rollouts, retaining only action-faithful ones for dense safety-aware scoring and scene-level closed-loop RL post-training.
Extensive experiments on nuScenes and an in-house dataset with over 130K training scenarios demonstrate consistent gains across planners, reducing safety violations by \emph{27.6\%} with DiffusionDrive on nuScenes and \emph{33.7\%} with Qwen3-VL on internal data.
The code is available at \href{https://github.com/Hongbin98/RoXDrive}{\textcolor{red}{\emph{RoXDrive}}}.

\end{abstract}

\section{Introduction}
End-to-end (e2e) autonomous driving aims to learn a direct mapping from raw sensory observations to future trajectories or low-level control commands~\citep{chen2024end}. 
Recent works have achieved remarkable progress in both model architectures~\citep{chitta2022transfuser,hu2023planning,jiang2023vad,liao2025diffusiondrive,jia2025drivetransformer,sun2025sparsedrive,li2026discrete,zheng2026driveagent,zhang2026resworld} and benchmarks~\citep{caesar2020nuscenes,dosovitskiy2017carla,dauner2024navsim}. 
Despite this progress, most e2e agents are still trained primarily through imitation learning (IL) from logged expert demonstrations. 
Such open-loop supervision is scalable, but it creates a fundamental mismatch between training and deployment: the policy never observes how its own actions change future scenes. 
Consequently, agents may exploit shortcut correlations, suffer from causal confusion, and generate trajectories that perform well under open-loop metrics yet degrade when executed under the closed-loop real-world deployment.

\begin{figure}[t]
    \centering
    \includegraphics[width=0.98\linewidth]{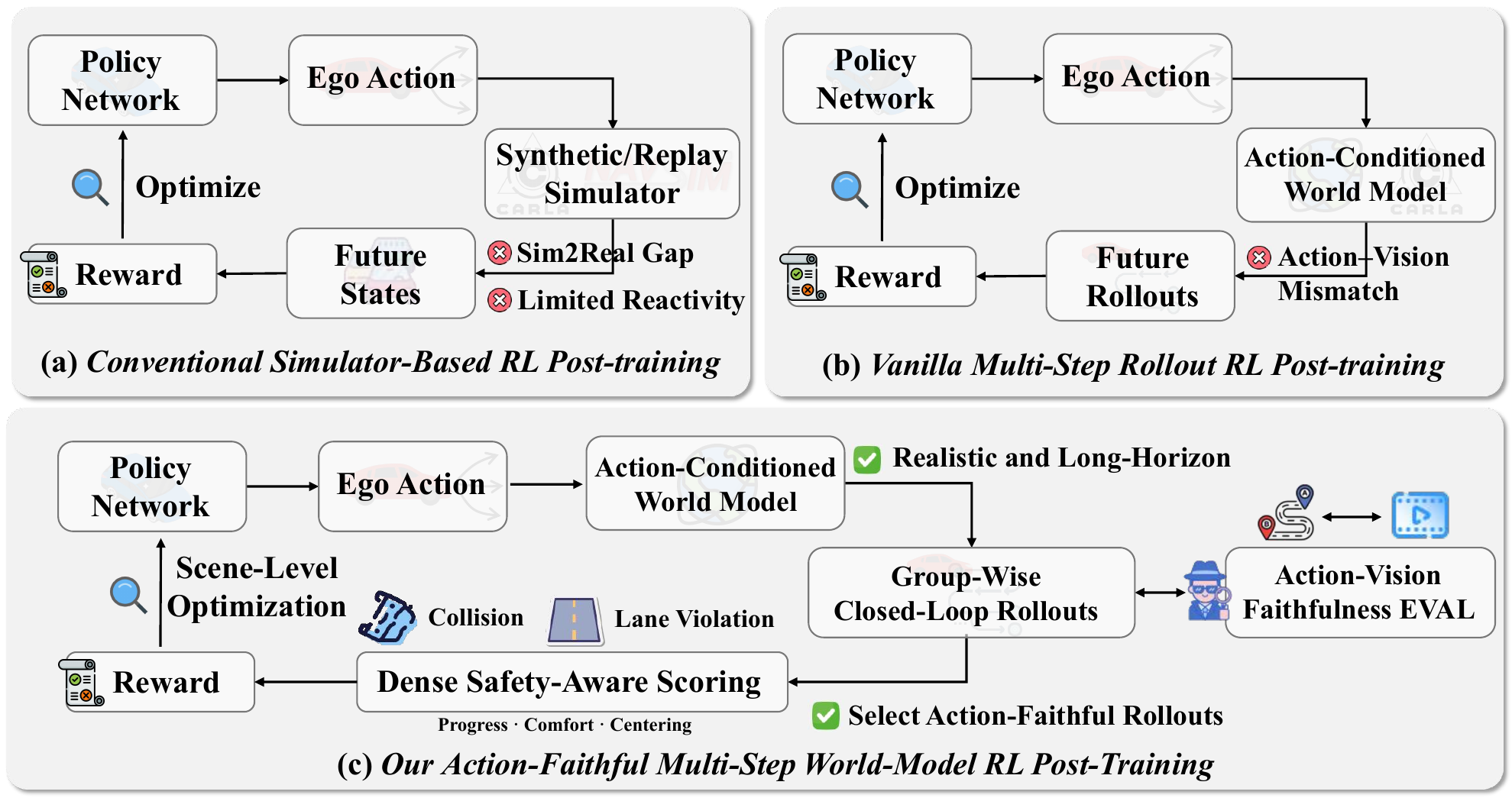} 
    \caption{(a) Conventional synthetic and reconstruction-based simulators trade off interactive freedom and real-scene fidelity.
    (b) Action-conditioned video world models enable realistic long-horizon interaction, but action–vision mismatch may produce unreliable consequences.
    (c) \ournet~explicitly measures action-vision faithfulness, retaining reliable episodes to optimize agents.
    }
    \label{fig:paradigm}
\end{figure}

Reinforcement learning (RL) provides a natural way to alleviate this issue by exposing agents to the future consequences of their actions through interactive environments, thereby improving decision-making~\citep{zhang2026minddriver,garcia2026road,shang2026drivedpo}.
However, the key difficulty lies in constructing suitable interactive environments. 
Real-world RL is costly and raises safety concerns, whereas synthetic simulators such as CARLA~\citep{dosovitskiy2017carla} provide controllable interaction~\citep{li2024think2drive,yang2025raw2drive} but exhibit sim-to-real gaps in visual appearance, traffic behavior, and scenario distribution.
Based on neural rendering or 3D Gaussian Splatting, reconstruction-based simulators are able to preserve real-scene appearance~\citep{gao2025rad,yan2026ad}, but are typically tied to recorded data and thus lack realistic counterfactual evolution like departures from recorded ego trajectories and reactive agent behaviors.
To summarize, existing environments offer either interactive freedom at the expense of realism, or real-scene fidelity with limited counterfactual interaction, as illustrated in Fig.~\ref{fig:paradigm} (a).

Recent action-conditioned video world models, such as GAIA-3~\citep{wayve2025gaia3}, Cosmos 3~\citep{agarwal2026cosmos} and X-World~\citep{zheng2026x} suggest promising alternatives by generating realistic and temporally consistent future driving scenes.
However, \emph{visual realism does not guarantee action-vision faithfulness}. During multi-step generation, visual dynamics may gradually deviate from conditioning ego actions, leading to accumulated action-vision inconsistencies and incorrect policy optimization as shown in Fig.~\ref{fig:paradigm}  (b).
It raises a key question: 
Can we explicitly evaluate action-vision faithfulness to quantify causal errors and identify reliable world-model rollouts, thereby enabling video world models to serve as reliable simulators for multi-step closed-loop policy optimization?

\begin{figure}[t]
    \centering
    \includegraphics[width=0.98\linewidth]{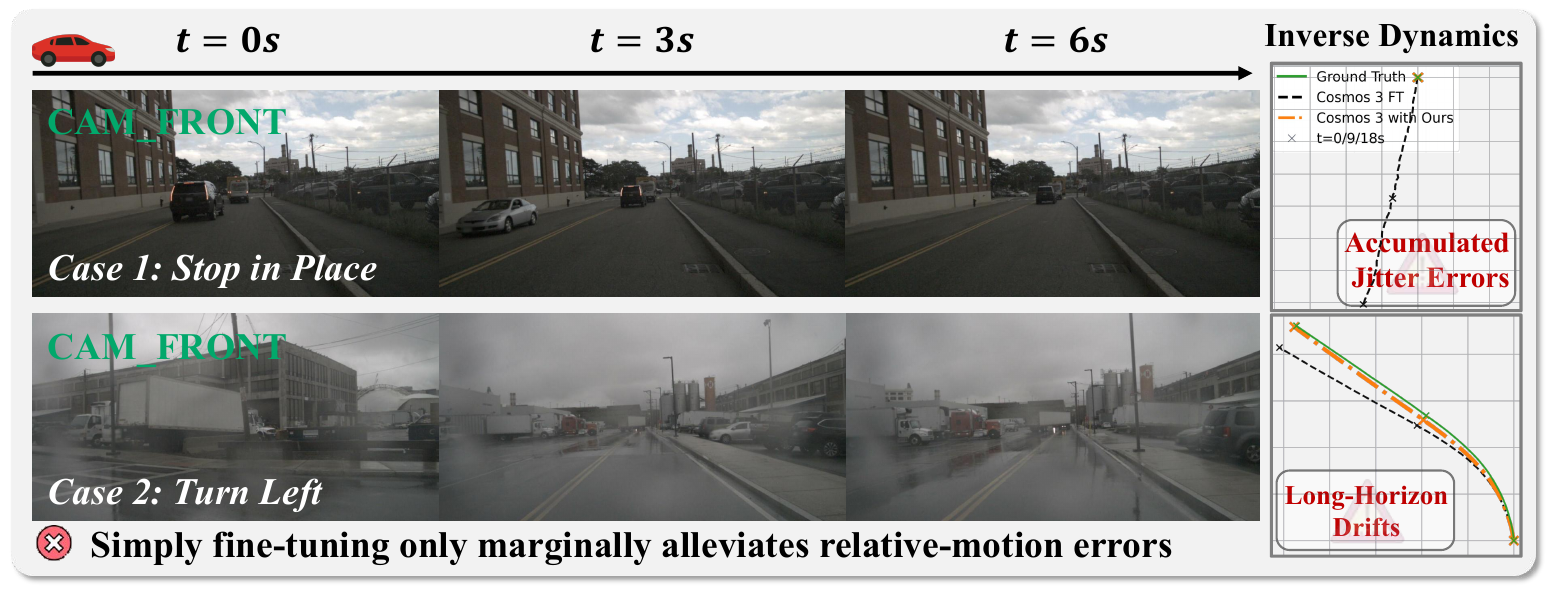} 
    \caption{
    Naive fine-tuning of powerful Cosmos 3 (\ie Cosmos 3 FT) remains insufficient for long-horizon inverse dynamics, \eg jitter errors (top) and long-horizon trajectory drifts (bottom). Hence, \ournet~devises geometry-aware auxiliary supervision to handle these issues.
    }
    \label{fig:motivation}
\end{figure}

To this end, we propose \ournet, a plug-and-play closed-loop RL framework with two stages as shown in Fig.~\ref{fig:paradigm}  (c): 
1) Model pre-training: 
We initialize the policy through imitation learning and develop an Action-Vision Faithfulness Evaluator (AVFE) which estimates ego motion from video and compares it with the conditioning actions to assess action-vision faithfulness.
However, directly fine-tuning Cosmos 3 suffers substantial cumulative relative-motion errors in long-horizon inverse dynamics estimation as shown in Fig.~\ref{fig:motivation}.
We therefore introduce the geometry-aware auxiliary trajectory supervision strategy, significantly mitigating the relative-motion errors.
2) Action-faithful RL post-training: 
Starting from an initial driving scene, the pre-trained policy iteratively interacts with a frozen action-conditioned video world model, acting on future observations induced by its own actions until the episode terminates or a failure occurs.
Then, AVFE filters these rollouts and retains only action-faithful ones for subsequent policy optimization.
To provide effective reinforcement signals, \ournet~further devises the \emph{Dense Safety-Aware Scoring} strategy which assigns fine-grained rewards to generated rollout states by evaluating collision clearance, lane clearance, ego progress, comfort, and lane centering.
We then apply Group Relative Policy Optimization (GRPO)~\citep{shao2024deepseekmath} at the scene level, comparing multiple rollouts within the same scene group to get relative advantages.
Unlike recent GRPO-based driving methods~\citep{zou2025diffusiondrivev2,jiang2025alphadrive}, our groups consist of long-horizon and reliable closed-loop episodes induced by the policy, enabling optimization from compounding long-horizon consequences.

Our contributions are summarized as follows: 
1) To the best of our knowledge, we are the first to explicitly evaluate action–vision faithfulness of long-horizon world-model rollouts in e2e driving. 
With our geometry-aware trajectory supervision, the action-vision faithfulness evaluator mitigates cumulative errors in inverse dynamics estimation and identifies reliable rollouts for policy optimization.
2) With reliable rollouts, \ournet~introduces the Dense Safety-Aware Scoring and Scene-level Closed-Loop GRPO strategies to provide stable and fine-grained reinforcement signals.
Video diffusion world models are only adopted during post-training, without additional inference cost at deployment.
3) We validate the proposed \ournet~on the public nuScenes~\citep{caesar2020nuscenes} and further conduct large-scale training and evaluation across real-world scenarios, including 130K challenging training scenarios and 1K test scenarios
of narrow roads, vulnerable road users, and dense multi-agent interactions, demonstrating consistent improvements in safety and driving quality.

    

\section{Related Work}
We first review the literature on world models in autonomous driving, and then discuss their usage for policies.
More discussions on related fields (\eg e2e autonomous driving) are put in Appendix~\ref{supp_e2e}.

\subsection{World Models in Autonomous Driving}
Conventional end-to-end driving paradigms such as diffusion-based~\citep{liao2025diffusiondrive}, regression-based~\citep{chitta2022transfuser}, and score-based approaches~\citep{li2025generalized} typically map current observations directly to control actions. 
While effective for reactive behavior, such formulations lack an explicit mechanism for anticipating future risks or reasoning about long-horizon consequences. World models address this issue by learning an internal dynamics model that can simulate plausible future evolutions of the driving scene conditioned on the current state. 
A prominent line of work focuses on {latent world models}~\citep{li2024enhancing,shi2025drivex,zhang2026resworld}, which predict future scene representations in a compact feature space instead of the raw pixel domain \citep{wang2024drivedreamer,wang2024driving}. 
These predicted latent states can then support downstream reasoning tasks, including trajectory scoring and selection~\citep{li2025end}, causal inference over potential outcomes~\citep{wang2026latent,li2025drivevla}, and chain-of-thought (CoT) deliberation~\citep{tan2025latent,zeng2025futuresightdrive}, thereby providing a predictive substrate for planning under uncertainty \citep{min2024driveworld}. 
Another family of approaches equips the model with explicit {visual–geometric foresight} through autoregressive generation of future tokens that are subsequently decoded into multi-task prediction targets (e.g., occupancy, depth, or semantic maps).
For example, DriveDreamer~\citep{zhou2026drivedreamer} jointly synthesizes future depth maps, video frames, and driving actions within a unified generative framework. 
Similarly, PWM~\citep{zhao2025forecasting} first imagines long-horizon future videos and then derives actions from the imagined rollouts. Furthermore, Uni-World VLA~\citep{liu2026uni} interleaves action generation with high-frequency future frame synthesis, allowing the policy to continuously update its anticipation and adapt its behavior within a coupled perception–action loop.

\subsection{Enhancing Policy Learning with World Models}

Pure imitation learning (IL) often suffers from {causal confusion}, where policies exploit spurious correlations in the training data rather than capturing the true causal drivers of expert behavior. 
A common strategy to mitigate this issue is to combine reinforcement learning (RL) \citep{yang2026worldrft} with IL, enabling policy refinement through trial-and-error optimization of an explicit reward signal. In autonomous driving, however, RL-based policy improvement critically relies on closed-loop interaction, where the agent executes rollouts in an environment that provides meaningful feedback. Existing interaction environments present several limitations. 
High-fidelity simulators often simplify the behavior of surrounding agents and suffer from a persistent sim-to-real gap~\citep{li2024think2drive,yang2025raw2drive}. 
Photorealistic reconstructions based on 3D Gaussian Splatting~\citep{kerbl20233d} offer high visual realism but incur prohibitive computational costs~\citep{gao2025rad}. 
Meanwhile, traffic-flow simulators, though computationally efficient, lack interactive 3D geometry and thus limit the transferability of learned policies~\citep{dauner2024navsim,li2025recogdrive,liu2026reinforced}. To overcome these challenges, recent work leverages learned world models as efficient and differentiable surrogates for closed-loop policy training.  AD-R1~\citep{yan2026ad} introduces an {Impartial Occupancy World Model} that predicts future occupancy grids and computes collision-based rewards directly in grid space, enabling scalable on-policy optimization without relying on external simulators. 
LaST-VLA~\citep{luo2026last} aligns action generation with COSMOS-based latent reasoning signals, optimizing trajectory-level rewards while using latent chain-of-thought features as stable internal guidance.
RAD-2~\citep{gao2026rad} proposes {BEV-Warp}, a high-throughput feature-level simulation environment that exploits spatial equivariance to accelerate policy iteration. 

\section{Preliminaries}
\label{sec:prel}
\textbf{Video Diffusion World Model.}
In autonomous driving, world models~\citep{ha2018world} are often devised to model the dynamics of the future scene, allowing e2e planners to learn how their actions shape the evolution of the future scene.
Building upon WAN~\citep{wan2025wan}, existing approaches~\citep{zheng2026x} adopt the latent video generation paradigm that couples a video spatio-temporal variational autoencoder with the mainstream DiT-based~\citep{peebles2023scalable} latent denoiser.
In this paper, we adopt the X-World as the action-conditioned multi-camera video diffusion world model which supports accurate scene control and causal generation.
Specifically, given recent synchronized multi-view videos $\mathbf{X}_{\mathrm{his}}$ and a future action sequence $\mathbf{a}$, the model predicts the resulting future camera observations $\hat{\mathbf{X}}_{\mathrm{fut}}$ via:
\begin{equation}
\hat{\mathbf{X}}_{\mathrm{fut}}
\sim
p_\omega\!\left( 
\mathbf{X}_{\mathrm{fut}} \mid
\mathbf{X}_{\mathrm{his}},
\mathbf{a},
\mathbf{c}
\right),
\end{equation}
where $\mathbf{c}$ denotes the optional scene control conditions, including dynamic agents, static elements, camera parameters, and scene descriptions. 
To support streaming inference, a chunk-wise causal generator rolls out future videos sequentially:
\begin{equation}
\hat{\mathbf{X}}_{\mathrm{fut}}^{(k)}
\sim
p_\omega\!\left(
\cdot \mid
\mathbf{H}^{(k)},
\mathbf{a}^{(k)},
\mathbf{c}
\right),
\quad
\mathbf{H}^{(k)}
=
\left[
\mathbf{X}_{\mathrm{his}},
\hat{\mathbf{X}}_{\mathrm{fut}}^{(<k)}
\right].
\end{equation}
Here, $\hat{\mathbf{X}}_{\mathrm{fut}}^{(k)}$ denotes the generated $k$-th future chunk corresponding to the action $\mathbf{a}^{(k)}$.
$\mathbf{H}^{(k)}$ is the causal history context composed of the observed history 
$\mathbf{X}_{\mathrm{his}}$ and previously generated chunks 
$\hat{\mathbf{X}}_{\mathrm{fut}}^{(<k)}$.
With the same history context, different action inputs lead to different future evolutions.
More details on the action-conditioned world model are put in the Appendix~\ref{supp_xworld}. 
\\
\textbf{Group Relative Policy Optimization.}
Proximal Policy Optimization (PPO)~\citep{schulman2017proximal} is a widely used policy-gradient algorithm that stabilizes the training process by limiting how large the new policy can deviate from the old policy in each update.
Given a state $\mathbf{s}$ and an action sequence $\mathbf{a}$ sampled from the old policy, PPO updates the current policy according to the probability ratio
$p(\theta)
=
\frac{
\pi_{\theta}(\mathbf{a}\mid\mathbf{s})
}{
\pi_{\theta_{\mathrm{old}}}(\mathbf{a}\mid\mathbf{s})
}$
and then clips this ratio to avoid unstable updates.
The update is weighted by an advantage estimate ${A}$, typically obtained from a learned value function.
However, as noted in the previous work~\citep{shao2024deepseekmath}, training an additional critic introduces substantial memory and computational cost, which becomes particularly burdensome in world-model-based RL where each policy update requires multiple costly scene rollouts.
Therefore, we adopt scene-level Group Relative Policy Optimization (GRPO), which removes the explicit value function and instead computes relative advantages from a group of action rollouts under the same driving context.

\begin{figure}[t]
    \centering
    \includegraphics[width=0.94\linewidth]{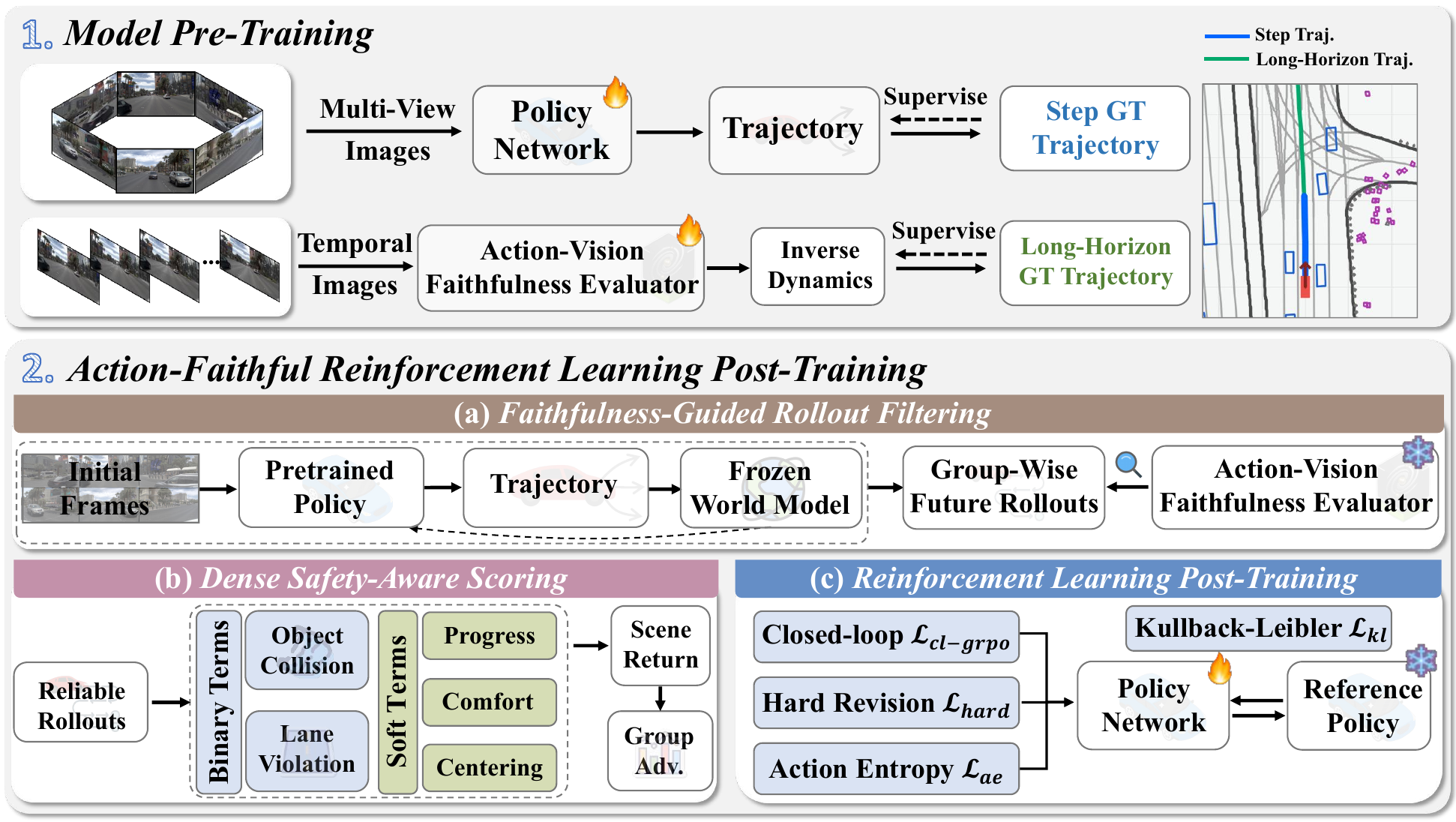} 
    \caption{Illustration of \ournet. 
    In stage one (top), we train the action-vision faithfulness evaluator with our geometry-aware auxiliary trajectory supervision beyond policy pre-training.
    In stage two (bottom), agents iteratively interact with a frozen world model to form long-horizon scene rollouts, retaining only action-faithful rollouts for dense safety-aware scoring and closed-loop RL.
    }
    \label{fig:method}
\end{figure}

\section{Closed-Loop Reinforcement Learning Post-Training}
\subsection{Method Overview}
As illustrated in Fig.~\ref{fig:method}, the overall framework contains two stages, \ie 
Model Pre-Training and Reinforcement Learning Post-Training.
Given a pre-trained planner in the standard open-loop manner on $\mathcal{D}$, we initialize the frozen reference policy $\pi_{\mathrm{ref}}$ and a trainable copy $\pi_{\theta}$.
For action-vision faithfulness, we further train an action-vision faithfulness evaluator $\mathcal{E}_{\phi}(\cdot)$ via inverse dynamics estimation with geometry-aware trajectory supervision $\mathcal{L}_{\mathrm{geo}}$ on temporal front-view image sequences:
\begin{equation}
\mathcal{L}_{\mathrm{AVFE}}
=
\mathcal{L}_{\mathrm{RF}}
+
\lambda_{\mathrm{geo}}m(\sigma)\mathcal{L}_{\mathrm{geo}},
\label{eqn:avfe_objective}
\end{equation}
where $\mathcal{L}_{\mathrm{RF}}$ denotes the inverse-dynamics Rectified Flow (RF) objective, and $\mathcal{L}_{\mathrm{geo}}$ provides geometry-aware supervision within informative RF noise levels, as detailed in Sec.~\ref{sec:avae}.

In the next stage, \ournet~optimizes $\pi_{\theta}$ through reinforcement learning post-training based on the frozen action-conditioned video diffusion world model $\mathcal{W}$ serving as the interactive environment.
Given an initial scene $\xi \sim \mathcal{D}$, \ournet~instantiates $G$ parallel world-model sessions from $\xi$.
At each decision step $l$ of rollout $g$, we denote the state as $\mathbf{s}_{l,g}=(\mathbf{o}_{l,g}, \mathbf{x}_{l,g})$, where $\mathbf{o}_{l,g}$ is the multi-view visual observation and $\mathbf{x}_{l,g}$ is the associated driving context.
The planner predicts $K$ trajectory candidates $\{\mathbf{a}_{l,g}^{k}\}_{k=1}^{K}$ with logits $\{{z}_{l,g}^{k}\}_{k=1}^{K}$, and samples the executed action (\ie a total of $n$ trajectory points) according to the softmax probabilities.
Conditioned on this action, $\mathcal{W}$ generates the next observation and updates the scene context, thereby exposing the planner to the future states induced by its own decisions.
All closed-loop rollouts continue until reaching the maximum horizon or a terminal safety violation, after which they are filtered by $\mathcal{E}_{\phi}(\cdot)$ to retain only faithful rollouts.

To leverage the reliable rollouts, \ournet~introduces the dense safety-aware scoring and closed-loop GRPO strategies.
We evaluate every $\mathbf{s}_{l,g}$ with safety and driving-quality terms which are accumulated into a long-horizon return for $G$ rollouts, including collision  $q^{\mathrm{obj}}$, lane violation $q^{\mathrm{lane}}$, ego progress $q^{\mathrm{prog}}$, comfort $q^{\mathrm{comf}}$, and lane centering $q^{\mathrm{ctr}}$.
Overall, \ournet~optimizes $\pi_\theta$ with:
\begin{equation}
\label{eq:total_loss}
    \theta^{\star}
    =
    \arg\min_{\theta}
    \Bigl\{
    \mathbb{E}_{\xi\sim\mathcal{D}}
    \left[
    \mathcal{L}_{\mathrm{cl\text{-}grpo}}(\xi)
    +
    \beta\mathcal{L}_{\mathrm{kl}}(\xi)
    -
    \lambda\mathcal{L}_{\mathrm{ae}}(\xi)
    \right]
    +
    \alpha\,
    \mathcal{L}_{\mathrm{hard}}
    \Bigr\}.
\end{equation}
where
$\mathcal{L}_{\mathrm{cl\text{-}grpo}}$ 
is calculated by faithful rollouts from $\mathcal{E}_{\phi}(\cdot)$,
$\mathcal{L}_{\mathrm{kl}}$ regularizes $\pi_\theta$ toward $\pi_{\mathrm{ref}}$,
$\mathcal{L}_{\mathrm{ae}}$ encourages diverse trajectory selection, and
$\mathcal{L}_{\mathrm{hard}}$ revisits hard scenes $\xi_h$, as detailed in Sec.~\ref{sec:rl}.
$\beta$, $\lambda$, and $\alpha$ are the weighting coefficients.
The pseudo-code of \ournet~is provided in Appendix~\ref{supp_pseudocode}.

\subsection{Action-Vision Faithfulness Evaluation}
\label{sec:avae}
To filter out mismatched world-model rollouts, we devise an action-vision faithfulness evaluator (AVFE) $\mathcal{E}_{\phi}(\cdot)$ based on Cosmos3-Nano~\citep{agarwal2026cosmos}, which estimates inverse dynamics from temporal front-view image sequences $\mathbf{I}$ as shown in Fig.~\ref{fig:method} (a).
Specifically, Cosmos 3 formulates action generation under the Rectified Flow (RF)~\citep{liu2022flow} objective, where inverse dynamics is realized by denoising action tokens $\mathbf{a}$ conditioned on $\mathbf{I}$ via: 
\begin{equation}
\label{eqn:rf_act}
\begin{gathered}
\mathbf{a}_{\sigma}
=(1-\sigma)\mathbf{a}
+\sigma\boldsymbol{\epsilon},
\quad
\mathbf{v}^{*}
=\boldsymbol{\epsilon}-\mathbf{a},
\\
\mathcal{L}_{\mathrm{RF}}
=
\left\|
v_{\phi}(\mathbf{a}_{\sigma},\sigma\mid\mathbf{I})
-\mathbf{v}^{*}
\right\|_{2}^{2}.
\end{gathered}
\end{equation}
Here $\boldsymbol{\epsilon}\sim\mathcal{N}(\mathbf{0},\mathbf{I}_d)$, $\sigma \in [0,1]$ denotes the noise level, $\mathbf{v}^{*}$ is the target RF velocity and $v_{\phi}(\cdot)$ represents the learnable velocity field of $\mathcal{E}_{\phi}$.
Each action token $\mathbf{a}_l$ represents relative ego-motions between adjacent frames, parameterized by 3D translation and 6D rotation.
Although this unified \emph{relative-pose} representation facilitates action modeling across heterogeneous embodiments, Eq.~\ref{eqn:rf_act} conducts token-wise supervision without explicitly constraining the composed trajectory.
Consequently, systematic translation and rotation errors may accumulate over time, leading to substantial drift in long-horizon estimation, \ie accumulated relative-motion errors shown in Fig.~\ref{fig:motivation}.

To this end, we introduce the geometry-aware auxiliary objective $\mathcal{L}_{\mathrm{geo}}$ for $\mathcal{E}_{\phi}$ via imposing trajectory-level supervision over the ego-motion trajectory.
Based on Eq.~\ref{eqn:rf_act}, we recover the clean action prediction at noise level $\sigma$ as $\hat{\mathbf{a}}_{\sigma}
=\mathbf{a}_{\sigma}-\sigma
v_{\phi}(\mathbf{a}_{\sigma},\sigma\mid\mathbf{I})$ while the ground-truth action satisfies $\mathbf{a}=\mathbf{a}_{\sigma}-\sigma\mathbf{v}^{*}$.
Thus, action prediction errors are related to RF velocity prediction errors:
\begin{equation}
\delta\mathbf{a}_{\sigma}
\triangleq
\hat{\mathbf{a}}_{\sigma}-\mathbf{a}
=
-\sigma
\left[
v_{\phi}(\mathbf{a}_{\sigma},\sigma\mid\mathbf{I})
-
\mathbf{v}^{*}
\right]
=
-\sigma\delta\mathbf{v},
\label{eqn:rf_action_error}
\end{equation}
where $\delta\mathbf{v}\triangleq
v_{\phi}(\mathbf{a}_{\sigma},\sigma\mid\mathbf{I})-\mathbf{v}^{*}$.
Eq.~\ref{eqn:rf_action_error} establishes a differentiable bridge from trajectory-level supervision on actions back to the original RF prediction.
To explicitly capture long-horizon error accumulation, we sequentially compose the predicted relative motions and define $\Phi_h(\cdot)$ as the differentiable composition up to horizon $h$.
Thus, the predicted $\hat{\boldsymbol{\tau}}_{h}$ and ground-truth $\boldsymbol{\tau}_{h}$ trajectory states are given by
$
\hat{\boldsymbol{\tau}}_{h}
=
\Phi_h(\hat{\mathbf{a}}_{\sigma,1:h}),
\boldsymbol{\tau}_{h}
=
\Phi_h(\mathbf{a}_{1:h}),
$
where $\boldsymbol{\tau}_{h}=(\mathbf{p}_h,\psi_h)$ contains the accumulated position and heading at horizon $h$.
Instead of supervising only the final state, we impose geometric constraints $\mathcal{L}_{\mathrm{geo}}$ at multiple predefined horizons $\mathcal{H}$ by:
\begin{equation}
\mathcal{L}_{\mathrm{geo}}
=
{\textstyle\sum_{h\in\mathcal{H}}}
\left[
\rho\!\left(\hat{\mathbf{p}}_h-\mathbf{p}_h\right)
+
\lambda_{\psi}
\rho\!\left(\hat{\psi}_h-\psi_h\right)
\right],
\label{eqn:traj_loss}
\end{equation}
where $\rho(\cdot)$ denotes the Smooth-$L_1$ loss and $\lambda_{\psi}$ balances position and heading supervision.
Let
$
\delta\boldsymbol{\tau}_{h}
=
\hat{\boldsymbol{\tau}}_{h}-\boldsymbol{\tau}_{h}
$
and
$
\mathbf{J}_{h}
=
\partial\boldsymbol{\tau}_{h}/\partial\mathbf{a}_{1:h}
$, a first-order approximation gives
$
\delta\boldsymbol{\tau}_{h}
\approx
\mathbf{J}_{h}\delta\mathbf{a}_{\sigma,1:h}
=
-\sigma\mathbf{J}_{h}\delta\mathbf{v}_{1:h}
$ combined with Eq.~\ref{eqn:rf_action_error}.
Hence, local RF errors are optimized based on their accumulated impact on the composed trajectory rather than solely on token-wise magnitudes.
Since the reliability of $\mathcal{L}_{\mathrm{geo}}$ varies across RF noise levels $\sigma$, we activate $\mathcal{L}_{\mathrm{geo}}$ only within a moderate noise range $m(\sigma)=\mathbb{I}\left[0.2\leq\sigma\leq0.7\right]$ which confines $\mathcal{L}_{\mathrm{geo}}$ to informative and stable RF regions.
Eventually, AVFE recovers the ego-action sequence for each rollout $g$ as $\hat{\mathbf{a}}_{g}
=\mathcal{E}_{\phi}(\mathbf{I}^{\mathcal{W}}_{g})$.
For the discrepancy of $\hat{\mathbf{a}}_{g}$ and $\mathbf{a}_g$, 
we normalize the error by $S_g
=
\max\left\{
\widetilde{\mathrm{ADE}}_g,\,
\widetilde{\mathrm{FDE}}_g,\,
\widetilde{e}^{\mathrm{yaw}}_g
\right\},$ including average displacement, final displacement, and mean yaw errors.
Therefore, only rollouts with $S_g\leq\eta$ are retained.
The pseudo-code and detailed analysis of AVFE are available in Appendix~\ref{supp_pseudocode} and~\ref{geo_analysis}, respectively.

\subsection{Scene-Level Closed-Loop Group Relative Policy Optimization}
\label{sec:rl}
Unlike open-loop GRPO methods, \ournet~performs group-relative optimization over policy-induced faithful rollouts by dense reward signals, sampling $G$ rollouts for each initial scene and obtaining all faithful rollouts $\mathcal{G}=\{g\in\{1,\ldots,G\}\mid S_g\leq\eta\}$ after AVFE filtering.\\
\noindent
\textbf{Dense Safety-Aware Scoring.}
For rollout $g$ at step $l$, the planner $\pi_{\theta}$ produces $K$ trajectory candidates with logits $\{\mathbf{a}_{l,g}^{k}, {z}_{l,g}^{k}\}_{k=1}^{K}$,  
where ${z}_{l,g}^{k}$ is the logit of the $k$-th trajectory in a fixed vocabulary.
The action index $k_{l,g}$ is sampled from $\mathrm{softmax}({z}_{l,g})$, yielding the executed trajectory $\mathbf{a}_{l,g}^{k_{l,g}}$.
Hence, $\mathcal{W}$ advances the closed-loop state by 
$\mathbf{s}_{l+1,g}
=
\mathcal{W}(\mathbf{s}_{l,g}, \mathbf{a}_{l,g}).
$
As shown in Fig.~\ref{fig:method} (b), 
to ensure safety, collision with dynamic objects (\eg cars) and crossing forbidden lanes (or road boundaries) are treated as terminal events with fixed negative costs $c_{\mathrm{fail}}$.
Given safe trajectories, \ournet~transforms $d^{\mathrm{obj}}$ and $d^{\mathrm{lane}}$, \ie the minimum clearance to nearby objects and non-crossable lanes, into dense scores by a piecewise function
$
\varphi(d;\delta^{-},\delta^{+})
=
\operatorname{clip}_{[0,1]}\!\left(
\frac{d-\delta^{-}}{\delta^{+}-\delta^{-}}
\right),
$
truncating to $[0,1]$.

Beyond safety, we consider progress, comfort, and centering terms to measure the quality of actions.
The progress term $q^{\mathrm{prog}}$ rewards the final displacement of the motion toward the expert future endpoint by $
q^{\mathrm{prog}}
=
\frac{e_{\mathrm{prog}}}{e_{\max}}
,$
where $e_{\mathrm{prog}}$ denotes the projected displacement toward the expert endpoint.
As for the comfort term, $q^{\mathrm{comf}}$ penalizes large speed variations along the trajectory by 
$q^{\mathrm{comf}}
=
1-
\frac{1}{n\Delta v_{\max}}
\sum_{i=1}^{n}
\left|v_{i}-v_{i-1}\right|.$
$q^{\mathrm{obj}}$ and $q^{\mathrm{lane}}$ are the object and lane clearance scores obtained by $\varphi(\cdot)$ while mutually exclusive lane centering term $q^{\mathrm{ctr}}$ prefers to balance the distance between the left $d^{l}$ and right sides $d^{r}$ by 
$q^{\mathrm{ctr}}=\frac{1}{n}
\sum_{i=1}^{n}
\left(1-\frac{|d^{l}_{i}-d^{r}_{i}|}
{d^{l}_{i}+d^{r}_{i}+\epsilon}
\right).$
Eventually, all terms yield normalized quality scores by truncating scores to $[0,1]$, including 
 $q^{\mathrm{prog}}$, $q^{\mathrm{comf}}$, and $q^{\mathrm{ctr}}$ (or $q^{\mathrm{obj}}$ and $q^{\mathrm{lane}}$).
The step reward aggregates all terms to obtain $r_{l,g}$ with negative costs for terminal violations:
\begin{equation}
r=\mathcal{S}(\mathbf{s},\mathbf{a})=
\begin{cases}
c_{\mathrm{fail}},
& \text{if } \mathbf{a} \text{ collides or hits forbidden lanes},\\
\bigl(\sum_{m\in\mathcal{M}} w_m q^{m}\bigr)
/
\bigl(\sum_{m\in\mathcal{M}} w_m\bigr),
& \text{otherwise}.
\end{cases}
\label{eq:dense_reward}
\end{equation}
Here $\mathcal{M}$ denotes valid terms and subscripts $(l,g)$ are omitted.
More details are put in Appendix~\ref{supp_reward}. \\
\textbf{Closed-Loop GRPO.}
Let $\mathcal{T}_g\subseteq\{0,\ldots,L-1\}$ denote the valid decision steps of rollout $g$ before termination.
The discounted return of rollout $g$ is $R_g = \sum_{l\in\mathcal{T}_g}\gamma^l r_{l,g}$.
Following GRPO~\citep{shao2024deepseekmath}, we normalize returns within the group to obtain the scene-relative advantage $A_g$ by:
\begin{equation}
A_g
=
\mathrm{clip}\bigl(
(R_g-\mu_R)/(\sigma_R+\epsilon),
-A_{\max}, A_{\max}
\bigr),
\label{eq:advantage}
\end{equation}
where the mean $\mu_R=\frac{1}{|\mathcal{G}|}\sum_{j\in\mathcal{G}}R_j$, and standard deviation $\sigma_R=\sqrt{\frac{1}{|\mathcal{G}|}\sum_{j\in\mathcal{G}}(R_j-\mu_R)^2}$.
We aggregate the log-probability of sampled actions along each rollout as
\begin{equation}
\bar{\ell}_g(\theta)
=
(1/|\mathcal{T}_g|)
\sum\nolimits_{l \in \mathcal{T}_g}
\log \pi_\theta(k_{l,g}\mid \mathbf{s}_{l,g}),
\label{eq:rollout_logprob}
\end{equation}
where $k_{l,g}$ is the sampled trajectory index at state $\mathbf{s}_{l,g}$. 
The closed-loop GRPO loss is then
\begin{equation}
\mathcal{L}_{\mathrm{cl\text{-}grpo}}
=
-(1/|\mathcal{G}|)
\sum\nolimits_{g\in\mathcal{G}}
A_g \bar{\ell}_g(\theta),
\label{eq:cl_grpo_loss}
\end{equation}
which increases the likelihood of long rollouts with higher relative returns within the same scene.\\
\textbf{Hard-Scene Revision.}
Since world-model interaction is computationally expensive, the group size $G$ is often limited and may fail to discover available distinct rollouts in challenging scenes. 
To this end, we revisit indistinguishable scenes with an expert-guided revision loss $\mathcal{L}_{\mathrm{hard}}$ by:
\begin{equation}
\textstyle
\mathcal{L}_{\mathrm{hard}}
=
\frac{1}{|\mathcal{D}_{\mathrm{hard}}|}
\sum_{\xi_h\in\mathcal{D}_{\mathrm{hard}}}
\Bigl[
-{z}_{h}^{k_h^{\mathrm{GT}}}
+\log\!\bigl(
\sum_{k=1}^{K}\exp({z}_{h}^{k})
\bigr)
+\bigl\|
\mathbf{a}_{h}^{k_h^{\mathrm{GT}}}
-\mathbf{a}_{h}^{\mathrm{GT}}
\bigr\|_1
\Bigr].
\label{eq:hard_revision_loss}
\end{equation}
where $k_h^{\mathrm{GT}}$ is the expert-matched trajectory index and $\mathbf{a}_{h}^{\mathrm{GT}}$ is the expert trajectory. 
Unlike RAD~\citep{gao2025rad}, $\mathcal{L}_{\mathrm{hard}}$ adaptively learns from hard scenes with insufficient RL signals.

\textbf{Regularization.}
As shown in Fig.~\ref{fig:method} (c), 
we further regularize policy with the action entropy term $\mathcal{L}_{\mathrm{ae}}$  to encourage exploration and action diversity while $\mathcal{L}_{\mathrm{kl}}$ constrains stable updates from $\pi_{\mathrm{ref}}$ via:
\begin{equation}
\mathcal{L}_{\mathrm{ae}}
=
\frac{1}{|\mathcal{G}|}
\sum_{g\in\mathcal{G}}
\frac{1}{|\mathcal{T}_g|}
\sum_{l \in \mathcal{T}_g}
\operatorname{Ent}
\left(
\pi_\theta(\cdot\mid \mathbf{s}_{l,g})
\right),
\quad
\mathcal{L}_{\mathrm{kl}}
=
\mathbb{E}_{l,g}
\left[
D_{\mathrm{KL}}
\left(
\pi_\theta(\cdot\mid \mathbf{s}_{l,g})
\|
\pi_{\mathrm{ref}}(\cdot\mid \mathbf{s}_{l,g})
\right)
\right].
\label{eq:regularization}
\end{equation}

\section{Experiments}
We conduct experiments to validate our method on the public nuScenes and a large-scale in-house dataset. Beyond merely imitating GT trajectories, our method aims to mitigate causal confusion as reflected in fewer collisions and off-road events.
Implementation details are put in the Appendix~\ref{supp_details}. \\
\noindent
\textbf{Datasets.}
The nuScenes~\citep{caesar2020nuscenes} dataset comprises 1K scenes with six camera views, each lasting 20 seconds, with keyframe annotations at 2 Hz. 
Following the protocol of MagicDriveV2~\citep{gao2025magicdrive}, we adopt 700 scenes for training and 150 for validation, and obtain the corresponding annotations at 12 Hz.
As for the in-house dataset, the training set contains over 130K scenarios, each lasting 30 seconds with annotations at 12 Hz and the held-out evaluation set includes 1K safety-critical scenarios, covering constrained road geometries, vulnerable road users, and dense interactions,
which are selected to assess safety and robustness under the closed-loop setting.
Note that OpenScene~\citep{openscene2023} (and its subset NAVSIM) is excluded due to its 2 Hz sampling rate while Cosmos~3, GAIA, and X-World are designed for more than 10 Hz.\\
\noindent\textbf{Compared Methods.}
We first evaluate \ournet~on representative e2e methods within nuScenes for fair comparisons, including TransFuser~\citep{chitta2022transfuser}, 
ST-P3~\citep{hu2022st}, UniAD~\citep{hu2023planning}, OccNet~\citep{tong2023scene}, VAD~\citep{jiang2023vad}, SparseDrive~\citep{sun2025sparsedrive}, DiffusionDrive~\citep{liao2025diffusiondrive} and VLAs, \ie Gemma-3~\citep{team2025gemma}, Qwen2.5-VL~\citep{bai2025qwen2.5vl} and Qwen3-VL~\citep{bai2025qwen3}.
Moreover, we also compare with e2e RL post-training methods, \ie CLEAR~\citep{shi2026clear} and Drive-r1~\citep{li2026drive}.
As for world models, we further provide action-following evaluation, including Cosmos3-Nano~\citep{agarwal2026cosmos}, Vista~\citep{gao2024vista}, Epona~\citep{zhang2025epona} and X-World~\citep{zheng2026x}.\\
\noindent\textbf{Evaluation Metrics.}
1) \emph{Object Collision}, the number of scenarios terminated by collision with objects;
2) \emph{Lane Violation}, the number of scenarios terminated by crossing a non-crossable lane boundary;
3) \emph{Progress}, the normalized forward progress;
4) \emph{Comfort}, the smoothness of velocity changes;
5) \emph{Clearance}, the risk of potential collisions and lane violations (replaced with \emph{Centering} on the in-house dataset).
Inspired by NAVSIM, we compute the final \emph{Driving Score} as $
\mathrm{DS}
=
\mathbb{I}_{\mathrm{safe}}
\cdot
\frac{
5{q}^{\mathrm{prog}}
+
3{q}^{\mathrm{comf}}
+
{q}^{\mathrm{obj}}
+
{q}^{\mathrm{lane}}
}{10},$
where $\mathbb{I}_{\mathrm{safe}}$ indicates no collision or lane violation.

\begin{table}[t]
\centering

\begin{minipage}[t]{0.44\linewidth}
\vspace{0pt}
\centering

\caption{
    Comparisons of inverse dynamics evaluation on the nuScenes validation set.
}
\label{tab:wm_idm_res}

\vspace{1mm}

\setlength{\tabcolsep}{2pt}
\renewcommand{\arraystretch}{0.92}

\resizebox{\linewidth}{!}{
\begin{tabular}{@{}l|ccc@{}}
    \toprule
    Method
    & 3s ADE $\downarrow$
    & 6s ADE $\downarrow$
    & 6s FDE $\downarrow$ \\
    \midrule

    Cosmos3-Nano
        & 8.77 & 15.07 & 30.11 \\

    \midrule
    GenAD$\ddagger$~\citep{yang2024generalized}
        & 0.90 & -- & -- \\
    Cosmos3-Nano Fine-Tuning
        & 0.67 & 1.27 & 2.67 \\
    \ournet~(Our AVFE)
        & \cellcolor{blue!8}\textbf{0.52}
        & \cellcolor{blue!8}\textbf{0.95}
        & \cellcolor{blue!8}\textbf{2.00} \\

    \midrule
    \multicolumn{4}{c}{With Strict SE(2)} \\
    \midrule

    Cosmos3-Nano
        & 8.82 & 15.36 & 31.27 \\

    \midrule
    Cosmos3-Nano Fine-Tuning
        & 0.66 & 1.24 & 2.59 \\
    \ournet~(Our AVFE)
        & \cellcolor{blue!8}\textbf{0.51}
        & \cellcolor{blue!8}\textbf{0.93}
        & \cellcolor{blue!8}\textbf{1.92} \\

    \bottomrule
\end{tabular}
}
\end{minipage}\hfill
\begin{minipage}[t]{0.53\linewidth}
\vspace{0pt}
\centering

\caption{
    Action-following evaluation on nuScenes over generated videos (\ie 6\,s, 72 frames) based on our AVFE.
    Bid./Cau. denote bidirectional/causal.
    Strict SE(2) retains $x$--$z$ translation and $y$-axis yaw.
}
\label{tab:idm_comparison}

\vspace{1mm}

\setlength{\tabcolsep}{2pt}
\renewcommand{\arraystretch}{1.08}

\resizebox{\linewidth}{!}{
\begin{tabular}{@{}l|cc|cccc@{}}
    \toprule
    Method
    & Bid./Cau.
    & \shortstack{Fine-\\Tuning}
    & \shortstack{SE(2) ADE\\(m) $\downarrow$}
    & \shortstack{SE(2) FDE\\(m) $\downarrow$}
    & \shortstack{Rotation\\Mean ($^\circ$) $\downarrow$}
    & \shortstack{Rotation\\Final ($^\circ$) $\downarrow$} \\
    \midrule

    GT Video
    & -- & --
    & 0.92 & 1.92 & 1.37 & 2.04 \\
    \midrule

    Cosmos3-Nano
    & Bid. & \xmark
    & 4.85 & 9.84 & 3.22 & 4.86 \\

    Vista
    & Cau. & \cmark
    & 4.19 & 8.03 & 4.63 & 7.34 \\

    Epona
    & Cau. & \cmark
    & 2.50 & 4.72 & 2.05 & 3.26 \\

    X-World
    & Cau. & \cmark
    & \cellcolor{blue!8}\textbf{1.11}
    & \cellcolor{blue!8}\textbf{2.18}
    & \cellcolor{blue!8}\textbf{1.77}
    & \cellcolor{blue!8}\textbf{2.94} \\

    \bottomrule
\end{tabular}
}
\end{minipage}

\end{table}

\vspace{-0.07in}
\begin{table}[t]
    \centering
    \caption{
        Comparison of open-loop (2s, collision) and closed-loop (4s) model performance on nuScenes.
        RL indicates reinforcement learning and $\ddagger$ denotes the official weights are unavailable.
    }
    \label{tab:nus_res}
    \vspace{1mm}
    
    \setlength{\tabcolsep}{4pt}
    \renewcommand{\arraystretch}{0.95}
    \resizebox{0.96\linewidth}{!}{
    \begin{tabular}{l c | lll | cc ccc c}
        \toprule
        \multirow{3}{*}{Method}
        & \multirow{3}{*}{RL}
        & \multicolumn{3}{c|}{Open-loop Col. (\%) $\downarrow$}
        & \multicolumn{6}{c}{Closed-loop Evaluation
          (4,675 4s clips, 2 action steps, based on X-World)} \\
        \cmidrule(lr){3-5}
        \cmidrule(lr){6-11}
    
        &
        & \multirow{2}{*}{1s}
        & \multirow{2}{*}{2s}
        & \multirow{2}{*}{Avg.}
        & \multicolumn{2}{c}{Safety}
        & \multicolumn{3}{c}{Driving Quality}
        & \multirow{2}{*}{Driving Score $\uparrow$} \\
        \cmidrule(lr){6-7}
        \cmidrule(lr){8-10}
    
        & & & &
        & Obj. Col. $\downarrow$
        & Lane Viol. $\downarrow$
        & Progress $\uparrow$
        & Comfort $\uparrow$
        & Clearance $\uparrow$
        & \\
        \midrule
    
        ST-P3
        & \xmark
        & 0.23 & 0.62 & 0.43
        & 268 & {557}
        & 0.397 & 0.425 & 0.387 & 0.341 \\
    
        VAD
        & \xmark
        & 0.07 & 0.17 & 0.12
        & 820 & 824
        & 0.255 & 0.681 & \textbf{0.418} & 0.271 \\
    
        UniAD
        & \xmark
        & 0.62 & 0.58 & 0.60
        & 348 & 870
        & 0.408 & 0.805 & {0.415} & 0.430 \\
    
        CLEAR$\ddagger$
        & \cmark
        & 0.11 & 0.23 & 0.17
        & -- & -- & -- & -- & -- & -- \\
    
        Drive-r1$\ddagger$
        & \cmark
        & 0.02 & 0.06 & 0.04
        & -- & -- & -- & -- & -- & -- \\
    
        \midrule
    
        DiffusionDrive
        & \xmark
        & 0.068 & 0.073 & 0.070
        & 266 & 772
        & {0.694} & 0.866 & 0.392 & 0.526 \\
    
        \textbf{+ \ournet~(Ours)}
        & \cmark
        & \cellcolor{blue!8}0.029 & \cellcolor{blue!8}0.063 & \cellcolor{blue!8}0.046
        & \cellcolor{blue!8}{{189}}
        & \cellcolor{blue!8}{\textbf{562}}
        & \cellcolor{blue!8}{0.717}
        & \cellcolor{blue!8}{0.883}
        & \cellcolor{blue!8}0.380
        & \cellcolor{blue!8}{0.588} \\

        SparseDrive
        & \xmark
        & \textbf{0.000} & 0.044 & 0.022
        & 197 & 623
        & 0.716 & 0.891 & 0.384 & 0.580 \\

        \textbf{+ \ournet~(Ours)}
        & \cmark
        &\cellcolor{blue!8}\textbf{0.000} &\cellcolor{blue!8}\textbf{0.015} & \cellcolor{blue!8}\textbf{0.007}
        & \cellcolor{blue!8}\textbf{173} & \cellcolor{blue!8}{598}
        & \cellcolor{blue!8}\textbf{0.775} & \cellcolor{blue!8}\textbf{0.922} & \cellcolor{blue!8}{0.388} & \cellcolor{blue!8}\textbf{0.622} \\
        
        \bottomrule
    \end{tabular}
    }

\end{table}

\begin{figure}[t]
    \centering
    \includegraphics[width=\linewidth]{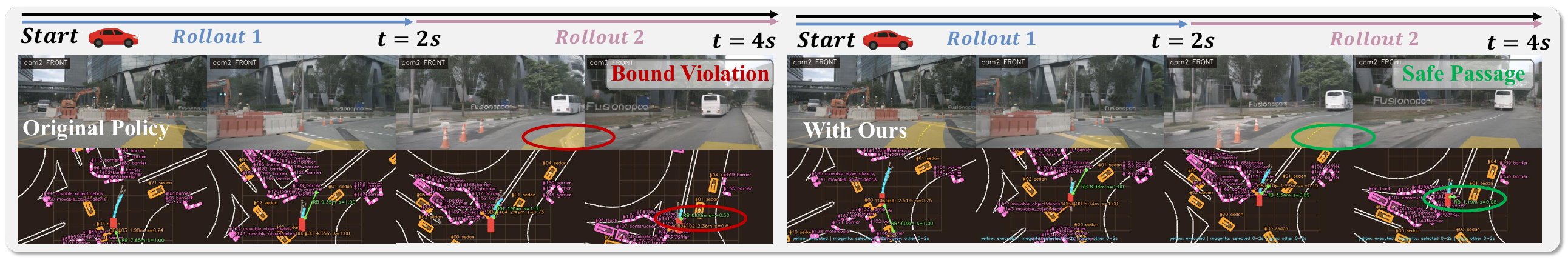}
    \vspace{-0.15in}
    \caption{Qualitative results of \ournet~avoiding risky behaviors (see also Appendix~\ref{supp_qua}).
    }
    \label{fig:vis}
\end{figure}

\begin{table}[t]
    \centering
    \caption{Closed-loop performance on 1K in-house test scenes (80 frames, 12 Hz) based on VLAs.}
    \label{tab:main_results}
    \setlength{\tabcolsep}{4pt}
    \renewcommand{\arraystretch}{0.85}
    \resizebox{\linewidth}{!}{
    \begin{tabular}{l l cc ccc c}
    \toprule
    \multirow{2}{*}{Planner (10 actions)} &
    \multirow{2}{*}{Training} &
    \multicolumn{2}{c}{Safety} &
    \multicolumn{3}{c}{Driving Quality} &
    \multirow{2}{*}{DS $\uparrow$} \\
    \cmidrule(lr){3-4}
    \cmidrule(lr){5-7}
    &
    &
    Obj. Col. $\downarrow$ &
    Lane Viol. $\downarrow$ &
    Progress $\uparrow$ &
    Comfort $\uparrow$ &
    Centering $\uparrow$ &
    \\
    \midrule
    

    

    

    \multirow{2}{*}{Gemma-3-4B}
    & Imitation-only
    & 328 & 171 & \textbf{0.938} & \textbf{0.952} & 0.521
    & 0.371 \\
    & \textbf{+ \ournet}
    & \cellcolor{blue!8}\textbf{198}
    & \cellcolor{blue!8}\textbf{127}
    & \cellcolor{blue!8}0.865
    & \cellcolor{blue!8}0.940
    & \cellcolor{blue!8}\textbf{0.539}
    & \cellcolor{blue!8}\textbf{0.540} \\

    \midrule

    \multirow{2}{*}{Qwen2.5-VL-3B}
    & Imitation-only
    & 331 & 141 & \textbf{0.945} & 0.911 & 0.538
    & 0.394 \\
    & \textbf{+ \ournet}
    & \cellcolor{blue!8}\textbf{206}
    & \cellcolor{blue!8}\textbf{107}
    & \cellcolor{blue!8}0.883
    & \cellcolor{blue!8}\textbf{0.936}
    & \cellcolor{blue!8}\textbf{0.547}
    & \cellcolor{blue!8}\textbf{0.559} \\

    \midrule

    \multirow{2}{*}{Qwen3-VL-2B}
    & Imitation-only
    & 242 & 162 & \textbf{0.884} & 0.938 & 0.527
    & 0.450 \\
    & \textbf{+ \ournet}
    & \cellcolor{blue!8}\textbf{174}
    & \cellcolor{blue!8}\textbf{94}
    & \cellcolor{blue!8}0.878
    & \cellcolor{blue!8}\textbf{0.957}
    & \cellcolor{blue!8}\textbf{0.542}
    & \cellcolor{blue!8}\textbf{0.608} \\
    
    \bottomrule
    \end{tabular}
    }
\end{table}

\begin{table}[!h]
    \centering
    \caption{Ablations of our AVFE and group size on nuScenes. More ablations are put in Appendix~\ref{supp_results}.}
    \label{tab:ablation_combined}
    \setlength{\tabcolsep}{3.5pt}
    \renewcommand{\arraystretch}{0.9}
    \resizebox{0.9\linewidth}{!}{
    \begin{tabular}{l|cccccc}
        \toprule
        Setting
        & Obj. Col. $\downarrow$
        & Lane Viol. $\downarrow$
        & Progress $\uparrow$
        & Comfort $\uparrow$
        & Clearance $\uparrow$
        & DS $\uparrow$ \\
        \midrule
        DiffusionDrive (official) & 266 & 772 & {0.694} & 0.866 & \textbf{0.392} & 0.526 \\
        {~$\bullet~$RL w. All Rollouts}
        & 237 & 672 & 0.701 & 0.881 & 0.385 & 0.562
        \\
        {~$\bullet~$RL w. Cosmos3-FT} 
        & 224 & 667 & 0.706 & 0.882 & 0.388 & 0.568 \\
        {~$\bullet~$RL w. AVFE (Ours)}
        & \cellcolor{blue!8}{\textbf{189}}
        & \cellcolor{blue!8}{\textbf{562}}
        & \cellcolor{blue!8}\textbf{0.717}
        & \cellcolor{blue!8}{\textbf{0.883}}
        & \cellcolor{blue!8}0.380
        & \cellcolor{blue!8}{\textbf{0.588}} \\

        \midrule
        \midrule
        Group Size & Obj. Col. $\downarrow$
        & Lane Viol. $\downarrow$
        & Progress $\uparrow$
        & Comfort $\uparrow$
        & Clearance $\uparrow$
        & DS $\uparrow$ \\
        \midrule
        DiffusionDrive (official) & 266 & 772 & {0.694} & 0.866 & \textbf{0.392} & 0.526 \\
        ~$\bullet~$$G=2$ & 246 & 636 & 0.679 & 0.857 & {0.383} & 0.552\\
        ~$\bullet~$$G=4$ & 198 & 614  & 0.714 & {0.884} & 0.382 & 0.580 \\
        ~$\bullet~$$G=6$ (default) 
        & \cellcolor{blue!8}{{189}}
        & \cellcolor{blue!8}{\textbf{562}}
        & \cellcolor{blue!8}{0.717}
        & \cellcolor{blue!8}{{0.883}}
        & \cellcolor{blue!8}0.380
        & \cellcolor{blue!8}{{0.588}}
        \\
        ~$\bullet~$$G=8$ & \textbf{180} & 588 & \textbf{0.720} & \textbf{0.890} & 0.381 & \textbf{0.590} \\
        \bottomrule
    \end{tabular}
    }
\end{table}

\subsection{Main Results}
\noindent
\textbf{Evaluation of AVFE and Action-Vision Faithfulness.} 
We first present the comparisons of inverse dynamics estimation on a total of 150 nuScenes validation scenes as shown in Table~\ref{tab:wm_idm_res}, demonstrating that our AVFE significantly mitigates the relative-motion errors instead of simply fine-tuning the powerful Cosmos 3, \eg reducing 3s Average Displacement Error (ADE) and 6s ADE by 22.4\% and 25.2\%.
Moreover, based on our AVFE, we assess action-following of world models on the same validation scenes as shown in Table~\ref{tab:idm_comparison}, revealing substantial differences in action-vision faithfulness across world models.
Combined with Table~\ref{tab:ablation_combined}, it highlights the importance of reliable optimization.
\\
\noindent
\textbf{Model Improvement with \ournet.}
Based on Table~\ref{tab:idm_comparison}, we adopt X-World as the interactive training environment to enable future scene generation.
On nuScenes, Table~\ref{tab:nus_res} shows that existing methods exhibit poor closed-loop performance even with only two action steps but achieve strong open-loop performance. 
Then, \ournet~improves DiffusionDrive and SparseDrive in both open-loop and closed-loop metrics, showing consistent gains in safety (\eg reducing failures by 27.6\% for DiffusionDrive) and driving scores.
To validate \ournet~at scale, we enhance large VLAs with \ournet~on the 130K in-house dataset and evaluate them on 1K test scenes as shown in Table~\ref{tab:main_results}, demonstrating consistent improvements in safety and DS across various architectures.
 

\subsection{Ablation Study}
We conduct ablation studies with DiffusionDrive on nuScenes to examine AVFE and the effect of group size as shown in Table~\ref{tab:ablation_combined}.
Compared with the official DiffusionDrive baseline, adding our RL strategy alone reduces unsafe cases by 12.4\%. 
Filtering rollouts with fine-tuned Cosmos 3 (Cosmos3-FT) further reduces 18 unsafe cases, whereas our AVFE reduces them by 158 (17.4\%) and achieves 0.588 DS. 
Regarding the group size, increasing $G$ from 2 to 4 substantially improves DS from 0.552 to 0.580.
Further increasing $G$ to larger values (\ie 6, 8) still yields notable performance gains, while we adopt $G=6$ as the default to balance performance and computational costs.

\subsection{Qualitative Results}
We provide qualitative visualizations of \ournet~in Figure~\ref{fig:vis}.
The results show that \ournet~enables DiffusionDrive to avoid risky behaviors, \ie keeping a safe distance from the road boundary, showing that existing planners can acquire safer behaviors without changing architectures by integrating \ournet.
More results are put in Appendix~\ref{supp_qua} and check our supp. for \emph{video demos}.

\section{Conclusion}
In this work, we present \ournet, a plug-and-play closed-loop RL framework that enables reliable policy optimization by leveraging action-faithful world-model rollouts.
With the geometry-aware auxiliary supervision, our action-vision faithfulness evaluator significantly mitigates cumulative errors in inverse dynamics estimation, enabling more reliable action-following evaluation of world models.
Based on reliable rollouts, \ournet~enables planners to learn from their own long-horizon action consequences with our dense safety-aware scoring and closed-loop GRPO strategies.
Experiments show consistent improvements in safety and overall driving scores, demonstrating the effectiveness of \ournet~in providing reliable closed-loop supervision for policy optimization. \\

\subsection*{AI use statement}
In this work, we utilize generative AI tools (\ie CodeX and ChatGPT) to improve our writing, assist with translation, refine the layout of the project page, and organize experimental results.
Specifically, we use these tools for the following purposes:
1) Notations and equations. We verify that all symbols are used consistently throughout the paper, ensuring all symbols and equations are clearly defined and explained.
2) Grammar and translation. We check and correct grammatical errors throughout the paper and check that translations accurately express our intention. 
3) Visualization. We reduce manual adjustments in the visualization process and obtain suggestions for improving the appearance of figures and the layout of the demos on the project page.
4) Organization. We utilize the AI tool to convert our results into LaTeX.
These tools were not used for other aspects, including method design and experiments. We have reviewed all AI-assisted content and take full responsibility for the final content of this work, including its text, claims, and accompanying artifacts.

\bibliography{iclr2027_conference}
\bibliographystyle{iclr2027_conference}

\newpage
\appendix

In the supplementary, we first provide more discussions on end-to-end autonomous driving.
Then, we provide more details of \ournet~and the usage of world models.
In addition, the pseudo-code of \ournet~is also provided.
Furthermore, we provide details and extended results to complement the main paper.
Our supplementary materials are organized as follows:

\begin{itemize}
    \item Section~\ref{supp_e2e} discusses existing end-to-end autonomous driving paradigms, including imitation learning and reinforcement learning.
    \item Section~\ref{supp_xworld} introduces how we leverage X-World as a closed-loop environment to generate multiple rollouts from the same initial scene.
    \item Section~\ref{supp_pseudocode} presents the pseudo-code of the proposed \ournet.
    \item Section~\ref{geo_analysis} gives more detailed analysis of the geometric constraint for our action-vision faithfulness evaluator and the ablation studies on the AVFE.
    \item Section~\ref{supp_reward} provides additional details of the reward terms.
    \item Section~\ref{supp_details} describes further implementation details and hyper-parameter settings.
    \item Section~\ref{supp_results} reports additional quantitative experimental results.
    \item Section~\ref{supp_qua} provides more qualitative visualizations.
    \item Section~\ref{supp_lim} provides more discussion on our limitations and future directions.
\end{itemize}

\section{More Discussions on E2E Autonomous Driving}
\label{supp_e2e}
Existing end-to-end autonomous driving methods can be broadly categorized into imitation learning (IL) and reinforcement learning (RL) paradigms~\citep{chen2024end}.
IL-based methods learn driving policies from expert demonstrations by minimizing the discrepancy between predicted actions and recorded human or expert trajectories.
Early approaches~\citep{pomerleau1988alvinn,bojarski2016end} directly mapped camera observations to control commands with neural networks.
Recent methods further improve planning performance by incorporating multi-sensor inputs, structured scene representations, or auxiliary supervision tasks~\citep{prakash2021multi,chitta2022transfuser,liao2025diffusiondrive}.
RL-based methods, in contrast, optimize policies according to reward signals that measure the expected quality of candidate actions under the current state~\citep{mnih2015humanlevel}.
This paradigm is appealing for autonomous driving because it can, in principle, account for long-term consequences beyond one-step imitation targets.
However, purely RL-based training remains challenging for complex driving systems, as reward signals are often sparse or noisy and the resulting gradients are insufficient to stably train large perception-planning architectures from scratch.
To improve training stability, several works introduce supervised objectives, auxiliary tasks, or imitation priors into RL pipelines~\citep{toromanoff2020end,chekroun2023gri}.

\section{More Discussions on Video Diffusion World Model}
\label{supp_xworld}
As mentioned before, OpenScene~\citep{openscene2023} (and its subset NAVSIM) is excluded due to its 2 Hz sampling rate and the latest Cosmos~3, GAIA, and X-World are designed for more than 10 Hz.
Therefore, we survey \emph{open-source} world models evaluated on nuScenes~\citep{caesar2020nuscenes} with 12 Hz sampling, including Cosmos3-Nano~\citep{agarwal2026cosmos}, Vista~\citep{gao2024vista}, and Epona~\citep{zhang2025epona}.
However, these models mainly support forward dynamics, \ie conditioning Image-to-Video generation on future ego trajectories, but do not provide explicit scene control over dynamic object states or the preservation of static scene geometry. Both capabilities are necessary for a world model to serve as a fair and controllable simulator for the training and evaluation of autonomous driving.
For this reason, we resort to the closed-source X-World~\citep{zheng2026x} as the world-model simulator in our experiments. X-World~\citep{zheng2026x} is an action-conditioned multi-camera video world model for autonomous driving with static scene geometry preservation. Given a synchronized multi-view visual history and a future ego-action sequence, it generates the corresponding future multi-camera observations in video space.
Its key property for our \ournet~is action causality: under the same initial scene, different ego actions can lead to different future observations while the generated videos remain temporally coherent and consistent with the commanded motion.

In our framework, X-World is sufficiently fine-tuned on the 700 training scenes of nuScenes to adapt it to the public dataset. Although this adaptation slightly compromises its inherent generative and generalization capabilities, it still produces reliable and realistic future observations. The original bidirectional model is converted to causal inference through Video-to-Video adaptation.
As for the in-house dataset, X-World is kept frozen and used only as an interaction environment.
For each sampled driving scene, we initialize multiple world-model sessions with the same visual history, ego state, surrounding agents, and static road context.
The planner then predicts a set of trajectories, from which different rollout workers sample or select different actions.
Each action is sent to an independent X-World session, which advances the scene and returns the next generated observation.
The planner observes this generated state again and repeats the process, forming a closed-loop rollout.
This design allows us to obtain multiple counterfactual futures from the same initial condition.
Since all rollouts share the same starting scene, their differences mainly come from the executed ego actions and the resulting world-model responses.
We therefore compare these rollouts within the same group and compute relative advantages from their long-horizon returns.
This \emph{same-initialization comparison is crucial for stable policy optimization}, since it removes much of the reward-scale variation caused by different scene difficulties and focuses learning on which action sequence leads to safer and more efficient closed-loop outcomes.

\section{Pseudo-code of \ournet}
\label{supp_pseudocode}

\begin{algorithm}[h]
\caption{The training pipeline of \ournet}
\label{alg:ournet}
\begin{algorithmic}[1]
\REQUIRE Dataset $\mathcal{D}$, policy $\pi_{\theta}$, world model $\mathcal{W}$, 
AVFE $\mathcal{E}_\phi$, group size $G$, rollout horizon $L$, multi-step $\mathcal{H}$, and
hyper-parameters $\beta,\lambda,\alpha,\lambda_{\mathrm{geo}}$
\STATE Pre-train $\pi_{\theta}$ on $\mathcal{D}$ in the open-loop manner
and obtain the frozen reference policy $\pi_{\mathrm{ref}}$;
\STATE Train $\mathcal{E}_\phi$ on $\mathcal{D}$ via Eq.~\ref{eqn:avfe_objective};
\STATE Initialize the hard-scene queue
$\mathcal{D}_{\mathrm{hard}}\gets\varnothing$;

\FOR{each training iteration}
    \STATE Sample a scene $\xi\sim\mathcal{D}$ and initialize $G$
    rollouts from the same initial state;

    \FOR{$t=0,\ldots,L-1$}
        \FOR{each active rollout $g$}
            \STATE Predict the candidate trajectories and their logits;
            \STATE Sample
            $k_{t,g}\sim\operatorname{softmax}(\mathbf{z}_{t,g})$
            and set
            $\mathbf{a}_{t,g}=\mathbf{a}_{t,g}^{k_{t,g}}$;
            \STATE Compute the step reward $r_{t,g}$ by
            Eq.~\ref{eq:dense_reward};
            \STATE Advance the closed-loop state by $\mathbf{s}_{t+1,g}=\mathcal{W}(\mathbf{s}_{t,g}, \mathbf{a}_{t,g})$; 
        \ENDFOR
    \ENDFOR

    \STATE Get $\mathcal{G}$ using $\mathcal{E}_\phi$ by Algorithm~\ref{alg:avfe};
    \STATE Compute the scene-level rollout returns $R_g$ and
    advantages $A_g$ by Eq.~\ref{eq:advantage};
    \IF{no valid rollout from $\xi$ satisfies the safety-progress criterion}
    \STATE Add $\xi$ to the hard-scene queue
    $\mathcal{D}_{\mathrm{hard}}$;
    \ENDIF
    
    \STATE Compute $\mathcal{L}_{\mathrm{cl\text{-}grpo}}$,
    $\mathcal{L}_{\mathrm{ae}}$, $\mathcal{L}_{\mathrm{hard}}$, and
    $\mathcal{L}_{\mathrm{kl}}$ by
    Eqs.~\ref{eq:cl_grpo_loss},
    \ref{eq:hard_revision_loss}, and
    \ref{eq:regularization};
    \STATE Update $\theta$ by minimizing the total objective in
    Eq.~\ref{eq:total_loss};
\ENDFOR

\RETURN The well-trained policy $\pi_{\theta^{\star}}$.
\end{algorithmic}
\end{algorithm}




\begin{algorithm}[h]
\caption{Rollout Filtering by Action-Vision Faithfulness via Our AVFE}
\label{alg:avfe}
\begin{algorithmic}[1]
\REQUIRE $G$ rollouts initialized from scene $\xi$, $\eta$, $\mathcal{E}_\phi$
\STATE Collect AVFE-eligible rollouts
$\mathcal{G}_{\mathrm{AVFE}}$;
\FOR{each rollout $g\in\mathcal{G}_{\mathrm{AVFE}}$}
    \STATE Get $\hat{\mathbf{a}}_{g} =\mathcal{E}_{\phi}(\mathbf{I}^{\mathcal{W}}_{g})$;
    \STATE Compute $S_g$ by Eq.\ref{eq:sr} and retain reliable rollouts by $S_g\leq\eta$;
\ENDFOR
\RETURN Reliable rollout group $\mathcal{G}$.
\end{algorithmic}
\end{algorithm}



\section{Analysis of Geometry-Aware Auxiliary Supervision}
\label{geo_analysis}
In this appendix, we further analyze how local Rectified Flow (RF) prediction errors propagate into long-horizon trajectory errors, and why our geometry-aware objective provides supervision beyond the token-wise RF objective.
Besides, we provide more ablation studies on the AVFE.

\begin{figure}[h]
    \centering
    \includegraphics[width=0.85\linewidth]{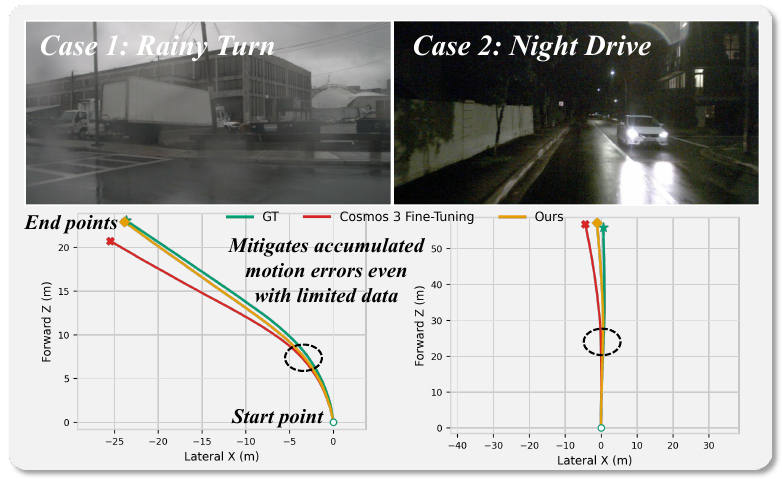}
    \caption{
        Qualitative inverse dynamics evaluation on representative nuScenes scenes.
    }
    \label{fig:idm_qualitative}
\end{figure}

As discussed in Section~\ref{sec:avae}, the effect of action errors on the composed trajectory is compactly characterized by the Jacobian $\mathbf{J}_{h}$.
Here, we make the geometric structure underlying this term explicit under planar SE(2) composition, where each ego motion is represented by a planar translation and a yaw rotation.
We show that translation and heading errors propagate differently through sequential composition. A local translation error directly perturbs the accumulated position, whereas a heading error additionally changes the orientation used to compose all subsequent translations.
Consequently, a heading error occurring earlier in the sequence affects a longer remaining portion of the trajectory and may induce larger positional drift. Such accumulated deviations are qualitatively shown in Figure~\ref{fig:idm_qualitative}, illustrating that local RF errors with similar token-wise magnitudes may lead to substantially different long-horizon trajectory errors after composition.

Let $(\mathbf{d}_{l},\theta_l)$ denote the planar translation and relative heading change obtained from the action token $\mathbf{a}_l$ after SE(2)
projection, and let $(\hat{\mathbf{d}}_{l},\hat{\theta}_l)$ denote their
predicted results.
We define the corresponding local errors as
$\delta\mathbf{d}_{l}=\hat{\mathbf{d}}_{l}-\mathbf{d}_{l}$
and
$\delta\theta_l=\hat{\theta}_l-\theta_l$.
The composed position and heading at horizon $h$ are given by: 
\begin{equation}
\mathbf{p}_{h}
=
\sum_{l=1}^{h}
\mathbf{R}(\psi_{l-1})\mathbf{d}_{l},
\qquad
\psi_h
=
\sum_{l=1}^{h}\theta_l,
\label{eq:supp_se2_composition}
\end{equation}
where $\psi_0=0$ and $\mathbf{R}(\psi)$ denotes the planar rotation matrix.
For the heading component, the accumulated error is equal to:
\begin{equation}
\delta\psi_h
=
\sum_{l=1}^{h}\delta\theta_l.
\label{eq:supp_yaw_accumulation}
\end{equation}
As for the position component, the heading error further changes the orientation
of subsequent translations
via the first-order approximation:
\begin{equation}
\mathbf{R}(\psi+\delta\psi)
\approx
\mathbf{R}(\psi)
+
\mathbf{R}(\psi)\mathbf{S}\delta\psi,
\qquad
\mathbf{S}
=
\begin{bmatrix}
0 & -1 \\
1 & 0
\end{bmatrix}.
\label{eq:supp_rotation_linearization}
\end{equation}
The accumulated position error can be approximated as:
\begin{equation}
\delta\mathbf{p}_{h}
\approx
\sum_{l=1}^{h}
\mathbf{R}(\psi_{l-1})\delta\mathbf{d}_{l}
+
\sum_{l=1}^{h-1}
\left[
\sum_{j=l+1}^{h}
\mathbf{R}(\psi_{j-1})
\mathbf{S}\mathbf{d}_{j}
\right]
\delta\theta_l.
\label{eq:supp_position_propagation}
\end{equation}
The first term describes the accumulated effect of \emph{local translation errors}, while
the second term shows that a heading error at step $l$ \emph{perturbs
every subsequent translation from step $l+1$ to $h$}.

Since $\mathbf{R}(\psi)$ and $\mathbf{S}$ preserve the Euclidean norm,
Eq.~\ref{eq:supp_position_propagation} further gives: 
\begin{equation}
\|\delta\mathbf{p}_{h}\|_2
\lesssim
\sum_{l=1}^{h}\|\delta\mathbf{d}_{l}\|_2
+
\sum_{l=1}^{h-1}
D_{l,h}|\delta\theta_l|,
\qquad
D_{l,h}
=
\sum_{j=l+1}^{h}\|\mathbf{d}_{j}\|_2.
\label{eq:supp_error_bound}
\end{equation}
Here, $D_{l,h}$ corresponds to the remaining travel distance after step $l$.
Therefore, \textbf{\emph{even for heading errors of the same magnitude, errors occurring
earlier in the sequence generally induce larger positional deviations}} because
they affect a longer portion of the subsequent trajectory.

Eq.~\ref{eqn:rf_action_error} illustrates that 
the reconstructed action error is proportional to the RF velocity error for a fixed noise level $\sigma$.
Therefore, minimizing $\mathcal{L}_{\mathrm{RF}}$ reduces local prediction errors in the action-token space.
As we mentioned above, it does not explicitly account for the geometric amplification introduced by sequential composition, \ie  Eq.~\ref{eq:supp_error_bound} shows that local errors with similar token-wise magnitudes can have different effects on the composed trajectory depending on their type and temporal location.
Moreover, errors with opposite signs may partially cancel, whereas small
systematic biases can accumulate coherently over time as shown in Figure~\ref{fig:motivation} (top).
Consequently, accurate token-wise reconstruction alone does not guarantee accurate long-horizon position and heading estimation.
To this end, for each supervised horizon $h$, we define: 
\begin{equation}
\ell_h
=\rho(\delta\mathbf{p}_h)+
\lambda_{\psi}\rho(\delta\psi_h),
\label{eq:supp_prefix_loss}
\end{equation}
here
$\mathcal{L}_{\mathrm{geo}}
=
\sum_{h\in\mathcal{H}}\ell_h$.
Based on the first-order eq. in Section~\ref{sec:avae}:
\begin{equation}
\delta\boldsymbol{\tau}_{h}
\approx
-\sigma
\mathbf{J}_{h}\delta\mathbf{v}_{1:h}.
\label{eq:supp_rf_traj_error}
\end{equation}
The trajectory-level correction propagated to the RF prediction satisfies:
\begin{equation}
\nabla_{\delta\mathbf{v}_{1:h}}\ell_h
\approx
-\sigma
\mathbf{J}_{h}^{\top}
\nabla_{\delta\boldsymbol{\tau}_{h}}\ell_h.
\label{eq:supp_geo_gradient}
\end{equation}
Eventually, the correction to the RF prediction is modulated by its geometric effect on the composed trajectory through $\mathbf{J}_{h}^{\top}$ rather
than being determined solely by its error magnitude in the original action space.
Particularly, error components that induce larger long-horizon position or heading deviations receive correspondingly stronger trajectory-level corrections.
Furthermore, we devise multiple-horizon supervision to prevent errors from being hidden by later compensation.
A loss considered only on the final state may remain limited when deviations at intermediate steps are partially canceled by subsequent motions.
In contrast, evaluating the composed state at multiple $h\in\mathcal{H}$ constrains intermediate prefixes as well as the final
trajectory state, thereby providing more direct supervision over the temporal evolution of accumulated errors. 

Furthermore, as mentioned in Section~\ref{sec:avae}, the RF objective is trained over the full sampled noise schedule, whereas the gate $m(\sigma)=\mathbb{I}\left[0.2\leq\sigma\leq0.7\right]$ applies only to $\mathcal{L}_{\mathrm{geo}}$.
Specifically, with a low $\sigma$, the geometry-aware auxiliary supervision becomes weak since its gradient with respect to the RF prediction is proportional to $\sigma$ (Eq.~\ref{eq:supp_geo_gradient}).
Besides, with a large $\sigma$, the heavily corrupted action makes the reconstructed trajectory unreliable. We therefore only consider the moderate range $\sigma_{\min}=0.2$ and $\sigma_{\max}=0.7$ for our geometry-aware objective.

\textbf{Ablation studies on the threshold $\eta$.}
For each rollout $g$, AVFE reconstructs the ego-action sequence as $\hat{\mathbf{a}}_{g} =\mathcal{E}_{\phi}(\mathbf{I}^{\mathcal{W}}_{g})$.
To measure the discrepancy, we compare the two trajectories (\ie $\hat{\mathbf{a}}_{g}$ and ${\mathbf{a}}_{g}$) at $T=48$ matched timestamps (\ie 12 Hz, 4 seconds) using Average Displacement Error, Final Displacement Error, and mean absolute yaw error $e_g^{\mathrm{yaw}}$ with angular differences wrapped to $[-\pi,\pi)$.
To achieve robust relative trajectory errors, we define the normalized discrepancy score by: 
\begin{equation}
S_g=\max\left\{
\frac{\mathrm{ADE}_g}{0.10\max(\overline L_g,1\,\mathrm{m})},
\frac{\mathrm{FDE}_g}{0.15\max(L_{g,T},1\,\mathrm{m})},
\frac{e_g^{\mathrm{yaw}}}{10^\circ}
\right\},
\label{eq:sr}
\end{equation}
where $L_{g,t}$ denotes the cumulative travel distance of the input trajectory up to step $t$, and
$\overline L_g=T^{-1}\sum_{t=1}^{T}L_{g,t}$.
For valid AVFE estimates, we retain rollouts with $S_g\leq\eta$, setting $\eta=0.75$ by default.
Here, $0.10$ and $0.15$ are fixed coefficients. Note that we skip the sample with $|\mathcal{G}|\leq1$.

Moreover, we present the ablations of $\eta$ under different settings with DiffusionDrive on nuScenes as shown in Table~\ref{tab:eta_abla}.
As $\eta$ increases to $0.75$, more rollouts with limited discrepancies are retained, progressively improving DS.
However, further increasing $\eta$ leads to a reduction in DS, suggesting that including less faithful rollouts introduces noise into RL post-training (\ie similar to \emph{RL with Cosmos3-FT} as shown in Table~\ref{tab:ablation_combined}).
We therefore adopt $\eta=0.75$ as the default, which achieves the highest DS and the fewest unsafe scenes, which further supports the effectiveness of rollout filtering.

\begin{table}[t]
    \centering
    \caption{Ablation studies of the filtering threshold $\eta$ on nuScenes.}
    \label{tab:eta_abla}
    \resizebox{0.85\linewidth}{!}{
    \begin{tabular}{lcccccc}
    \toprule
    Setting & Obj. Col.$\downarrow$ & Lane Viol.$\downarrow$
    & Progress$\uparrow$ & Comfort$\uparrow$
    & Clearance$\uparrow$ & DS$\uparrow$ \\
    \midrule
    DiffusionDrive
    & 266 & 772 & 0.694 & 0.866 & \textbf{0.392} & 0.526 \\
    $\eta=0.25$ & 209 & 624 & 0.691 & 0.869 & 0.382 & 0.565\\
    $\eta=0.50$
    & 205 & 630 & 0.707 & \textbf{0.883} & 0.384 & 0.575 \\
    $\eta=0.75$ (default)
    & \cellcolor{blue!8}{189}
    & \cellcolor{blue!8}{\textbf{562}}
    & \cellcolor{blue!8}\textbf{0.717}
    & \cellcolor{blue!8}\textbf{0.883}
    & \cellcolor{blue!8}{0.380}
    & \cellcolor{blue!8}\textbf{0.588} \\
    $\eta=1.25$
    & \textbf{184} & 587 & \textbf{0.717} & 0.882 & 0.380 & 0.585 \\
    $\eta=1.50$
    & 211 & 627 & 0.709 & 0.884 & 0.381 & 0.576 \\
    \bottomrule
    \end{tabular}
    }
\end{table}

\section{More Details of Reward Terms}
\label{supp_reward}
We calculate all reward terms in the ego-centric coordinate system for the rollout trajectory at each closed-loop decision step.
For safety evaluation, we check the motion against surrounding dynamic objects and non-crossable lane boundaries.
If the ego trajectory collides with a dynamic object or crosses a forbidden lane boundary, the rollout receives a terminal penalty with $c_{\mathrm{fail}}=-5.0$ and is stopped immediately.
Otherwise, we compute dense safety scores from the minimum object clearance $d^{\mathrm{obj}}$ and lane-boundary clearance $d^{\mathrm{lane}}$.
In our implementation, both clearance terms use the danger threshold $\delta^{-}=1.5\,\mathrm{m}$ and the safe threshold $\delta^{+}=3.0\,\mathrm{m}$.
For driving quality, we consider progress and comfort. The progress score measures the displacement toward the expert future
endpoint, while the comfort score penalizes large speed changes along the trajectory with $\Delta v_{\max}=2.5\,\mathrm{m/s}$.
The final reward is a weighted average of progress, comfort, and clearance (replaced with centering on the in-house dataset) with weights $5$, $3$, and $2$, respectively.
Each non-terminal transition receives a dense reward, while terminal safety failures provide immediate negative feedback.

For nuScenes, the ego vehicle is represented by an oriented bounding box of size $4.084\,\mathrm{m}\times1.85\,\mathrm{m}$, whose center is shifted forward by $0.5\,\mathrm{m}$ along the executed ego heading, following DiffusionDrive~\citep{liao2025diffusiondrive} and SparseDrive~\citep{sun2025sparsedrive}. A collision is detected when the filled ego polygon intersects an object polygon, where physical contact is also regarded as a collision. Cars and trucks are checked at all 12 Hz rollout frames~\citep{gao2025magicdrive}, while the remaining valid nuScenes object categories are additionally included at their exact annotated keyframes. The object clearance $d^{\mathrm{obj}}$ is the minimum polygon-to-polygon distance over the executed trajectory.
For lane-boundary checking, we use non-crossable road-boundary polylines provided by the nuScenes map annotations. A boundary violation occurs when the filled ego polygon intersects a non-crossable boundary polyline, and the lane-boundary clearance
$d^{\mathrm{lane}}$ is their minimum polygon-to-polyline distance.
Consequently, both object collisions and boundary violations have zero geometric clearance.
Since the official nuScenes only provides 2 Hz annotations of keyframes, we therefore replace the centering term with clearance scores (\ie there exists label noise of lanes~\citep{gao2025magicdrive}).

For the in-house dataset, we retain the original collision and non-crossable lane-boundary violation criteria.
A collision or boundary violation results in the same terminal penalty $c_{\mathrm{fail}}=-5.0$. Unlike the fixed ego footprint used for nuScenes, the oriented vehicle polygons are constructed according to the physical dimensions of the corresponding vehicles.
Note that clearance is not included as an independent reward term.
Instead, the clearance scores are used to gate the raw progress score by $
q^{\mathrm{prog}}
=
\frac{e_{\mathrm{prog}}}{e_{\max}}
\cdot
\min\left(q^{\mathrm{obj}}, q^{\mathrm{lane}}\right)
$. 
In addition, the centering term is evaluated only when valid lane boundaries are located within $12\,\mathrm{m}$.
When the centering term is unavailable, it is omitted and the remaining valid terms are re-normalized.

\section{More Implementation Details}
\label{supp_details}
We implement our method and all baselines in PyTorch~\citep{paszke2019pytorch}. All experiments are conducted on 6 nodes with 8 H800 GPUs each.
For AVFE training, we set $\lambda_{\mathrm{geo}}=10^{-3}$ and $\lambda_{\psi}=5.0$.
Geometry supervision is activated for $\sigma\in[0.2,0.7]$ at $\mathcal{H}=\{0.5,1.0,2.0,4.0,6.0\}\,\mathrm{s}$.
We use Smooth-$L_1$ parameters $\beta_p=1.0\,\mathrm{m}$ and $\beta_\psi=0.1\,\mathrm{rad}$, $e_{\max}=10\,\mathrm{m}$, and $\epsilon=10^{-12}$. 
Based on $\eta=0.75$, AVFE is applied only when a rollout contains a world-model observation subsequently consumed by the policy and its ego speed remains above $0.6\,\mathrm{m/s}$.
Otherwise, ego speeds at or below the above threshold are set to zero to suppress low-speed jitter.
We independently sample one RF noise level for each action sequence from a shifted logit-normal schedule with a shift of $10$ without additional high-noise sampling.

For nuScenes, all baselines are verified under the same and fair pipeline based on their official checkpoints. 
We use a 48-frame horizon consisting of two consecutive 24-frame rollouts.
At each stage, the planner enumerates six first-stage trajectory candidates, and each non-terminal candidate is further expanded into six second-stage candidates, forming a logical rollout group (up to $6\times6=36$) of two-stage trajectories per scene.
The global scene batch size is set to $4$ and the group size is set to $G=6$.
We set $\gamma=0.99$, $\beta=0.4$, $\alpha=1.0$, $\lambda=10^{-3}$, and the advantage clipping threshold to $A_{\max}=3.0$.
The sampling temperature is set to $1.0$.
We use AdamW with an initial learning rate of $10^{-4}$ and a weight decay of $0.01$. The learning rate is reduced to $3\times10^{-5}$ after 50 updates and to $10^{-5}$ after 100 updates.
The gradient clipping norms are set to $10.0$ for reinforcement learning and $30.0$ for hard scene revision.
Hard scene revision is performed every four updates with a hard scene queue size of $4$.
We optimize the classification and regression layers using the base learning rate and the remaining trainable decoder layers using $0.1$ times the base learning rate.

For the in-house dataset, all planners are initialized from supervised imitation-learning checkpoints, with the default 20K iterations and batch size 32.
During reinforcement post-training, the original imitation-trained planner is frozen as the reference policy $\pi_{\mathrm{ref}}$, while an additional trainable copy is optimized by \ournet.
X-World~\citep{zheng2026x} is also kept frozen and is used only as the closed-loop interaction environment.
The global scene batch size is set to $10$.
For each sampled scene, we initialize $G=4$ parallel X-World sessions from the same initial state.
Each rollout worker samples a trajectory candidate from the current planner and sends the selected action to a world-model session.
The generated future observation is then fed back to the planner for the next decision step.
Unless otherwise specified, each scenario contains 105 frames, and we perform closed-loop rollout for up to 80 future frames.
For policy optimization, we use AdamW as the optimizer and set the KL coefficient to $\beta=0.01$, entropy coefficient to $\lambda=10^{-3}$, hard-scene revision weight to $\alpha=1.0$, advantage clipping threshold to $A_{\max}=3.0$, sampling temperature to $1.0$, gradient clipping norm to $1.0$, and learning rate to $10^{-4}$. We set the maximum number of training iterations to 2K, \ie up to 20K scenes for closed-loop model training.
Besides, we maintain an online hard-scene queue for scenes in which all sampled rollouts fail to satisfy the safety-progress criterion.
These hard scenes are replayed with expert supervision to prevent the policy from drifting into unrecoverable unsafe modes.
The hard-scene queue size is set to 4, and at most 10 steps per hard scene are used for revision.
As for multi-modal planning, we add and fine-tune an anchor-selection head that predicts a probability distribution over 3,394 trajectory anchors of RAD~\citep{gao2025rad} and an offset regression head for trajectory refinement.
We train the modified models to convergence on the large-scale training data under the open-loop setting.

\begin{figure}[h]
    \centering
    \includegraphics[width=\linewidth]{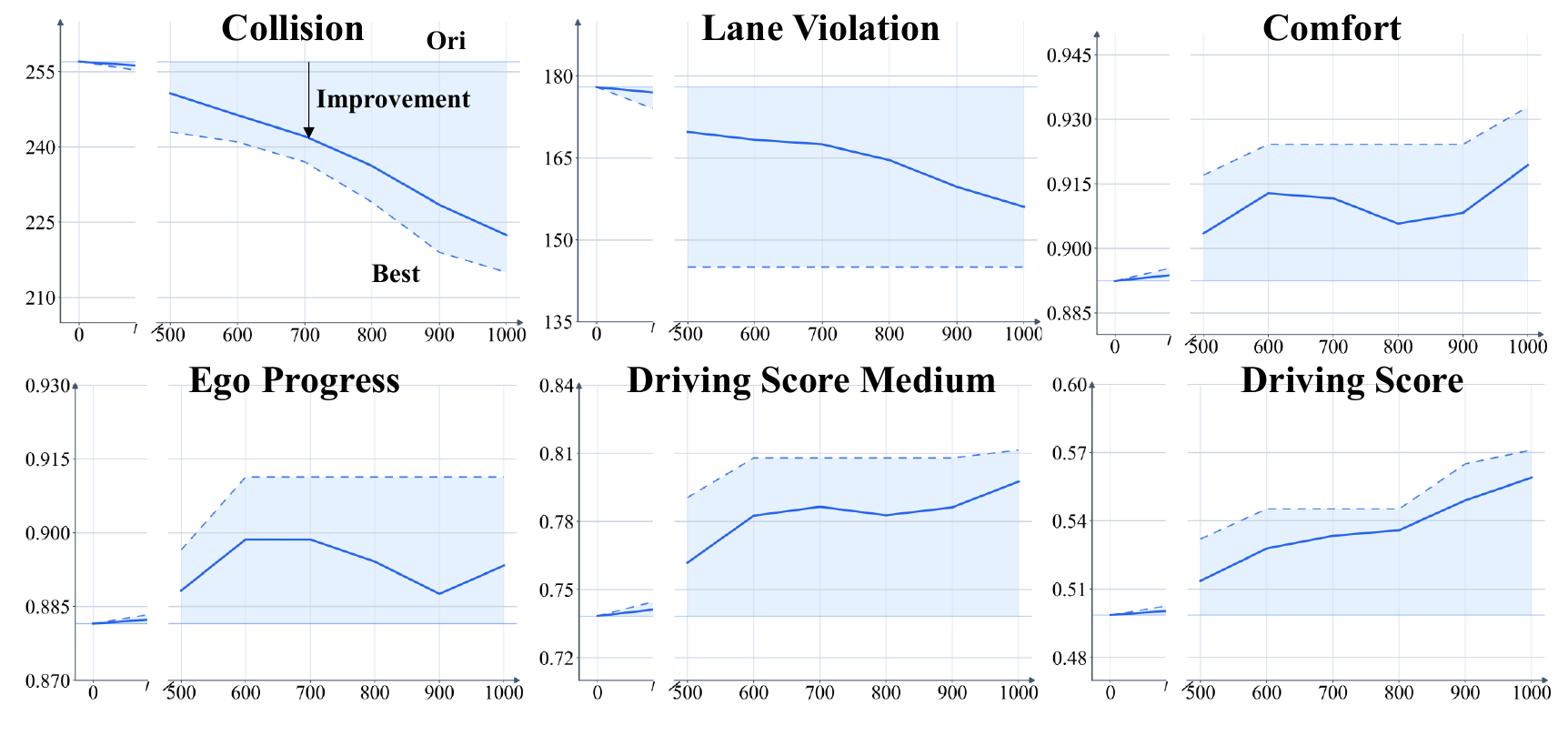}
    \vspace{-0.2in}
    \caption{Quantitative trends on test scenes during RL based on TransFuser.
    }
    \label{fig:ana1}
\end{figure}

\begin{table}[t]
    \centering
    \caption{Open-loop model performance on 1K test scenes in terms of ADE and FDE.}
    \label{tab:open_loop_metrics}

    \resizebox{0.8\linewidth}{!}{
    \begin{tabular}{@{}l|c c c c@{}}
        \toprule
        Model
        & ADE Median $\downarrow$
        & ADE Mean $\downarrow$
        & FDE Median $\downarrow$
        & FDE Mean $\downarrow$ \\
        \midrule
        TransFuser~\citep{chitta2022transfuser}
        & 0.060 & 0.075 & 0.116 & 0.146 \\
        GTRS~\citep{li2025generalized}
        & 0.052 & 0.049 & 0.105 & 0.104 \\
        \bottomrule
    \end{tabular}
    }
\end{table}

\begin{table}[t]
\centering
    \caption{Closed-loop model performance on 1K test scenes with a default frame horizon of 80.}
    \label{tab:closed_loop_backbone_results}
    \resizebox{\linewidth}{!}{
    \begin{tabular}{l|c c c c c|c}
    \toprule
    Model & Collision Scene $\downarrow$ & Lane Violation $\downarrow$ & Ego Progress $\uparrow$ & Comfort $\uparrow$ & Centering $\uparrow$ & Driving Score $\uparrow$ \\
    \midrule
    TransFuser~\citep{chitta2022transfuser} & 257 & 178 & 0.881 & 0.892 & \textbf{0.540} & 0.498 \\
    \textbf{\textbf{+ \ournet}} (Ours) & \cellcolor{blue!8}\textbf{215} & \cellcolor{blue!8}\textbf{145} & \cellcolor{blue!8}\textbf{0.900} & \cellcolor{blue!8}\textbf{0.932} & \cellcolor{blue!8}0.539 & \cellcolor{blue!8}\textbf{0.571} \\
    \midrule
    GTRS~\citep{li2025generalized} & 264 & 164 & \textbf{0.930} & 0.917 & 0.532 & 0.522 \\
    \textbf{\textbf{+ \ournet}} (Ours) & \cellcolor{blue!8}\textbf{214} & \cellcolor{blue!8}\textbf{126} & \cellcolor{blue!8}0.890 & \cellcolor{blue!8}\textbf{0.967} & \cellcolor{blue!8}\textbf{0.551} & \cellcolor{blue!8}\textbf{0.607} \\
    \bottomrule
    \end{tabular}
    }
\end{table}

\section{More Experimental Results}
\label{supp_results}

\noindent
\textbf{Training at scale with long-horizon sequences.}
1) Table~\ref{tab:open_loop_metrics} shows that both baselines are well pre-trained under the standard open-loop setting, achieving low Average Displacement Errors (ADE) and Final Displacement Errors (FDE) on the 1K test scenes.
2) However, open-loop imitation accuracy is not sufficient to ensure robust closed-loop performance.
As shown in Table~\ref{tab:closed_loop_backbone_results}, both imitation-only baselines suffer from severely unsafe terminations when verified in closed-loop rollouts, leading to unsatisfactory closed-loop metrics.
3) By applying \ournet, closed-loop safety and overall driving quality are consistently improved, \ie the total failure count is reduced by 17.2\% on TransFuser and 20.6\% on GTRS, revealing that our method provides consistent gains across various architectures.
Meanwhile, \ournet~calibrates ego progress to a more reasonable range and substantially improves comfort, while maintaining stable centering without biasing the planner toward aggressive modes.
4) Figure~\ref{fig:ana1} further illustrates the training process of \ournet.
Starting from the original policy, the right segment reports the performance of RL models (interval is set to 100 steps). 
The shaded regions indicate the improvements, showing that \ournet~progressively reduces safety failures while improving overall driving quality.
We further report the closed-loop performance of GTRS under various maximum frame horizons to illustrate the effectiveness of the proposed method, as shown in Table~\ref{tab:gtrs_mf_results}.

\begin{table}[h]
    \centering
    \caption{Closed-loop performance of GTRS under different maximum frame horizons (MF).}
    \label{tab:gtrs_mf_results}
    \resizebox{\linewidth}{!}{
    \begin{tabular}{l c|c c c c c|c}
    \toprule
    Model & MF & Object Collision & Lane Violation
    & Progress & Comfort & Centering & Driving Score \\
    \midrule
    GTRS~\citep{li2025generalized} & 20 & 31 & 19
    & \textbf{0.968} & 0.918 & 0.564 & 0.859 \\
    \textbf{+ \ournet} &
    & \cellcolor{blue!8}{\textbf{31}}
    & \cellcolor{blue!8}{\textbf{17}}
    & \cellcolor{blue!8}{0.944}
    & \cellcolor{blue!8}{\textbf{0.966}}
    & \cellcolor{blue!8}{\textbf{0.565}}
    & \cellcolor{blue!8}{\textbf{0.866}} \\
    \midrule
    GTRS & 40 & 73 & 50
    & \textbf{0.952} & 0.918 & 0.553 & 0.791 \\
    \textbf{+ \ournet} &
    & \cellcolor{blue!8}{\textbf{60}}
    & \cellcolor{blue!8}{\textbf{36}}
    & \cellcolor{blue!8}{0.922}
    & \cellcolor{blue!8}{\textbf{0.966}}
    & \cellcolor{blue!8}{\textbf{0.556}}
    & \cellcolor{blue!8}{\textbf{0.814}} \\
    \midrule
    GTRS & 60 & 202 & 123
    & \textbf{0.938} & 0.917 & 0.539 & 0.615 \\
    \textbf{+ \ournet} &
    & \cellcolor{blue!8}{\textbf{160}}
    & \cellcolor{blue!8}{\textbf{97}}
    & \cellcolor{blue!8}{0.901}
    & \cellcolor{blue!8}{\textbf{0.967}}
    & \cellcolor{blue!8}{\textbf{0.555}}
    & \cellcolor{blue!8}{\textbf{0.682}} \\
    \midrule
    GTRS & 80 & 264 & 164
    & \textbf{0.930} & 0.917 & 0.532 & 0.522 \\
    \textbf{+ \ournet} &
    & \cellcolor{blue!8}{\textbf{214}}
    & \cellcolor{blue!8}{\textbf{126}}
    & \cellcolor{blue!8}{0.890}
    & \cellcolor{blue!8}{\textbf{0.967}}
    & \cellcolor{blue!8}{\textbf{0.551}}
    & \cellcolor{blue!8}{\textbf{0.607}} \\
    \bottomrule
    \end{tabular}
    }
\end{table}

\noindent
\textbf{More ablation studies.}
To examine reward components and horizon lengths, we present the results under different settings based on TransFuser as shown in Table~\ref{tab:ablation_combined_supp}.
Compared with imitation-only, adding driving-quality terms leads to aggressive progress but causes more failures, leading to a lower DS.
By adding penalty terms, the planner obviously improves safety and calibrates ego progress, whereas using the full reward yields the best overall performance. 
Besides, \ournet~consistently improves model performance across all horizons with larger gains in longer rollouts (\ie more difficult), which further demonstrates our effectiveness.

 \begin{table}[t]
    \centering
    \caption{
    Ablation studies on reward components and the horizon lengths on the 1K test scenes.
    }
    \label{tab:ablation_combined_supp}


        \resizebox{\linewidth}{!}{
    \begin{tabular}{c c c c | c c | c c c | c}
        \toprule

        \multicolumn{10}{c}{(a) Reward components $R_{\mathrm{col}}$, $R_{\mathrm{lane}}$, and $R_{\mathrm{dq}}$ denote the collision, lane violation, and driving-quality terms.} \\
        \midrule

        TransFuser
        & $R_{\mathrm{col}}$
        & $R_{\mathrm{lane}}$
        & $R_{\mathrm{dq}}$
        & Obj. Col. $\downarrow$
        & Lane Viol. $\downarrow$
        & Progress $\uparrow$
        & Comfort $\uparrow$
        & Centering $\uparrow$
        & DS $\uparrow$ \\
        \midrule

        $\checkmark$ & $\times$ & $\times$ & $\times$
        & 257 & 178 & 0.881 & 0.892 & \textbf{0.540} & 0.498 \\

        $\checkmark$ & $\times$ & $\times$ & $\checkmark$
        & 372 & 185 & \textbf{0.943} & 0.860 & 0.529 & 0.396 \\

        $\checkmark$ & $\checkmark$ & $\times$ & $\checkmark$
        & 241 & 164 & 0.911 & 0.924 & 0.537 & 0.545 \\

        $\checkmark$ & $\times$ & $\checkmark$ & $\checkmark$
        & 229 & 156 & 0.889 & 0.899 & 0.539 & 0.539 \\

        $\checkmark$ & $\checkmark$ & $\checkmark$ & $\checkmark$
        & \cellcolor{blue!8}\textbf{215}
        & \cellcolor{blue!8}\textbf{145}
        & \cellcolor{blue!8}0.900
        & \cellcolor{blue!8}\textbf{0.932}
        & \cellcolor{blue!8}\textbf{0.539}
        & \cellcolor{blue!8}\textbf{0.571} \\

        \midrule
        \multicolumn{10}{c}{(b) Horizon lengths, at 12 Hz, 8 points per action} \\
        \midrule

        \multicolumn{2}{c}{Actions}
        & \multicolumn{2}{c|}{Model}
        & Obj. Col. $\downarrow$
        & Lane Viol. $\downarrow$
        & Progress $\uparrow$
        & Comfort $\uparrow$
        & Centering $\uparrow$
        & DS $\uparrow$ \\

        \midrule
        \multicolumn{2}{c}{\multirow{2}{*}{5 (3.33s)}}
        & \multicolumn{2}{c|}{TransFuser}
        & 74 & 55 & 0.915 & 0.892 & 0.550 & 0.768 \\
        \multicolumn{2}{c}{}
        & \multicolumn{2}{c|}{\textbf{+ \ournet}}
        & \cellcolor{blue!8}\textbf{63}
        & \cellcolor{blue!8}\textbf{46}
        & \cellcolor{blue!8}\textbf{0.927}
        & \cellcolor{blue!8}\textbf{0.932}
        & \cellcolor{blue!8}\textbf{0.553}
        & \cellcolor{blue!8}\textbf{0.795} \\        
        
        \midrule
        \multicolumn{2}{c}{\multirow{2}{*}{8 (5.33s)}}
        & \multicolumn{2}{c|}{TransFuser}
        & 199 & 126 & 0.892 & 0.893 & 0.543 & 0.593 \\

        \multicolumn{2}{c}{}
        & \multicolumn{2}{c|}{\textbf{+ \ournet}}
        & \cellcolor{blue!8}\textbf{163}
        & \cellcolor{blue!8}\textbf{104}
        & \cellcolor{blue!8}\textbf{0.910}
        & \cellcolor{blue!8}\textbf{0.933}
        & \cellcolor{blue!8}\textbf{0.544}
        & \cellcolor{blue!8}\textbf{0.659} \\

        \midrule

        \multicolumn{2}{c}{\multirow{2}{*}{10 (6.67s)}}
        & \multicolumn{2}{c|}{TransFuser}
        & 257 & 178 & 0.882 & 0.892 & \textbf{0.541} & 0.499 \\

        \multicolumn{2}{c}{}
        & \multicolumn{2}{c|}{\textbf{+ \ournet}}
        & \cellcolor{blue!8}\textbf{215}
        & \cellcolor{blue!8}\textbf{145}
        & \cellcolor{blue!8}\textbf{0.900}
        & \cellcolor{blue!8}\textbf{0.933}
        & \cellcolor{blue!8}0.540
        & \cellcolor{blue!8}\textbf{0.571} \\

        \bottomrule
    \end{tabular}
    }
\end{table}

\section{More Qualitative Results}
\label{supp_qua}
\begin{figure}[h]
    \centering
    \includegraphics[width=\linewidth]{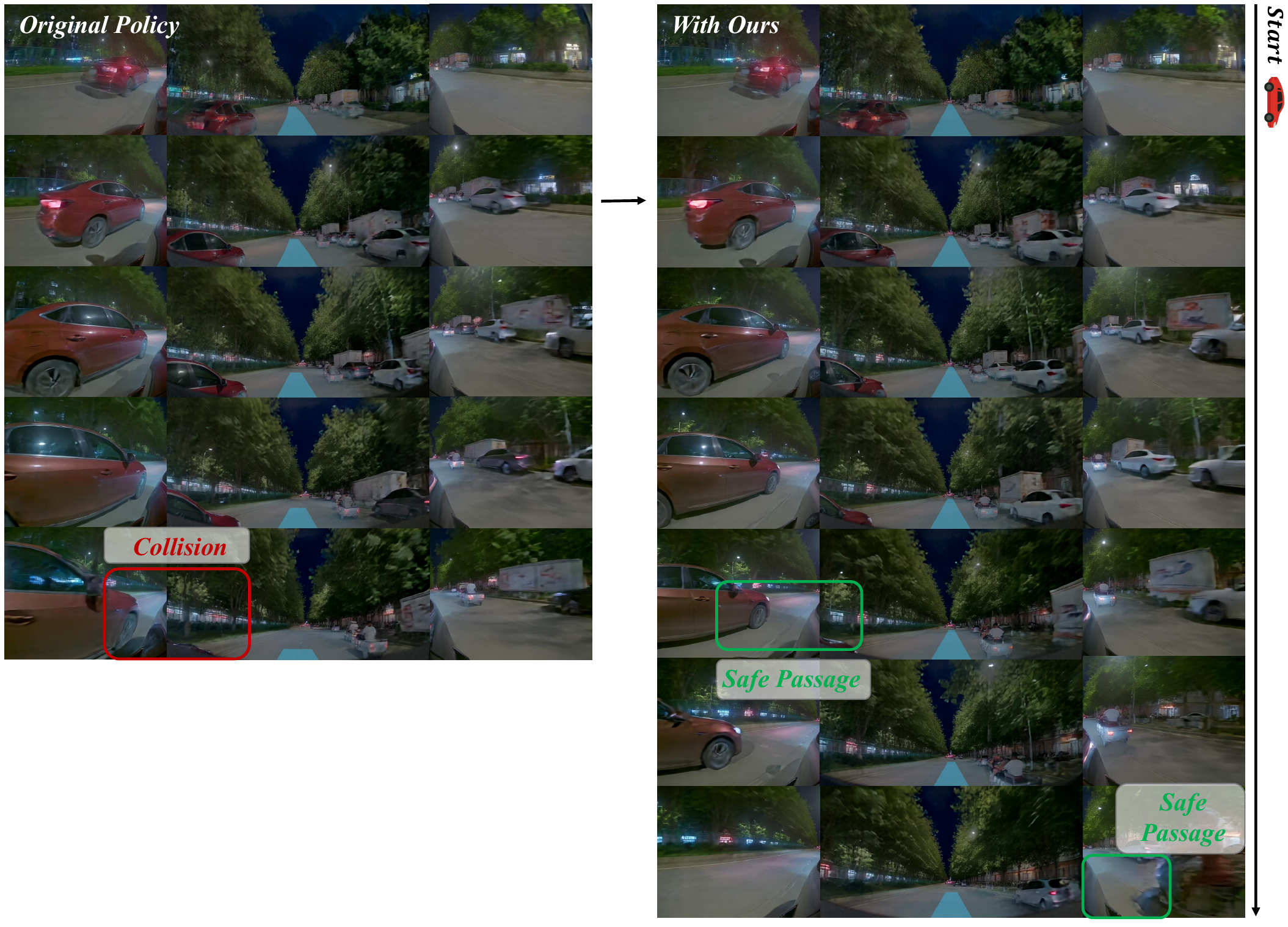} 
    \caption{Qualitative results of \ournet~avoiding the collision on the in-house dataset.}
    \label{fig:supp_vis1}
\end{figure}

\begin{figure}[h]
    \centering
    \includegraphics[width=\linewidth]{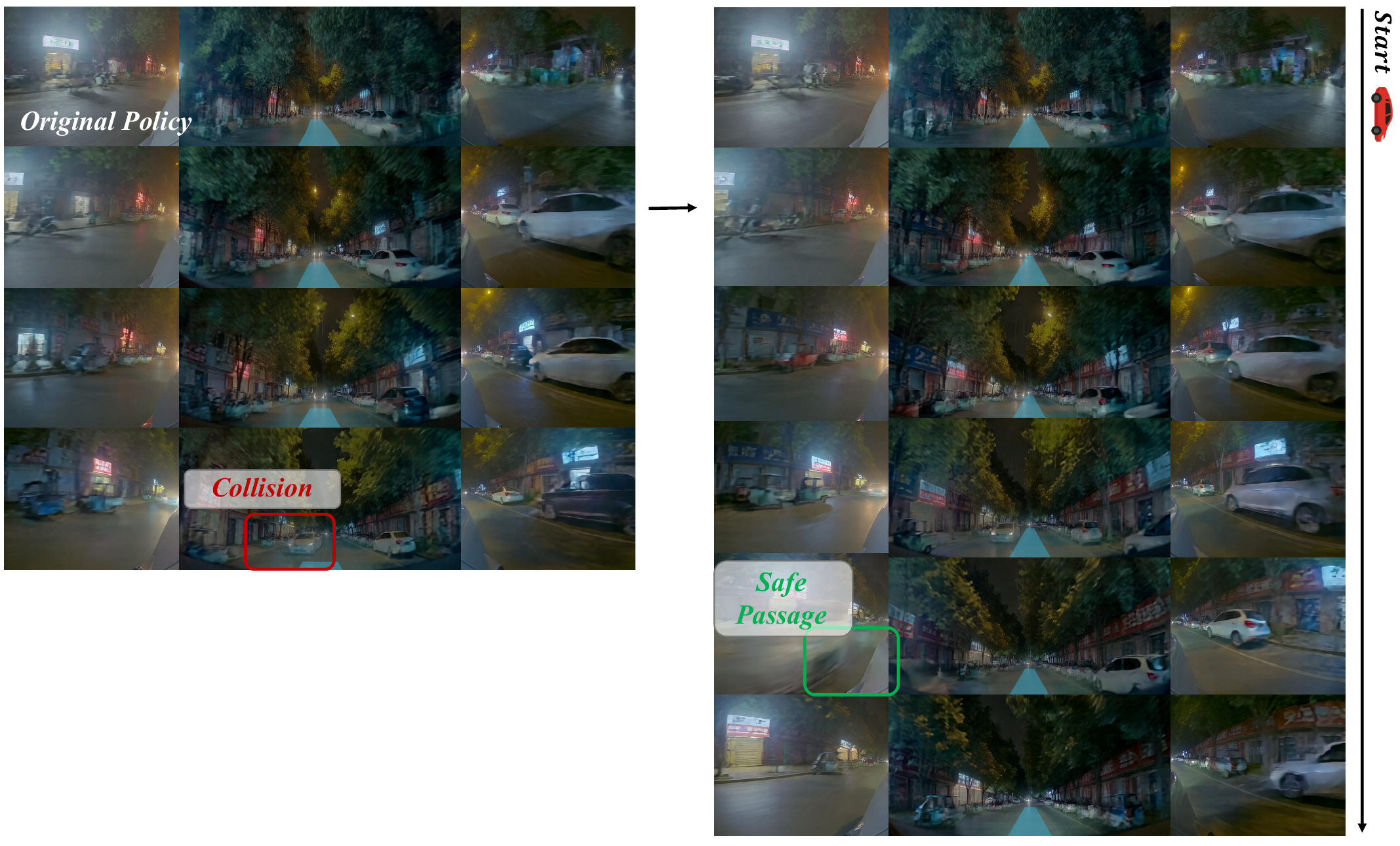} 
    \caption{Qualitative results of \ournet~avoiding the collision on the in-house dataset.}
    \label{fig:supp_vis3}
\end{figure}

\begin{figure}[h]
    \centering
    \includegraphics[width=\linewidth]{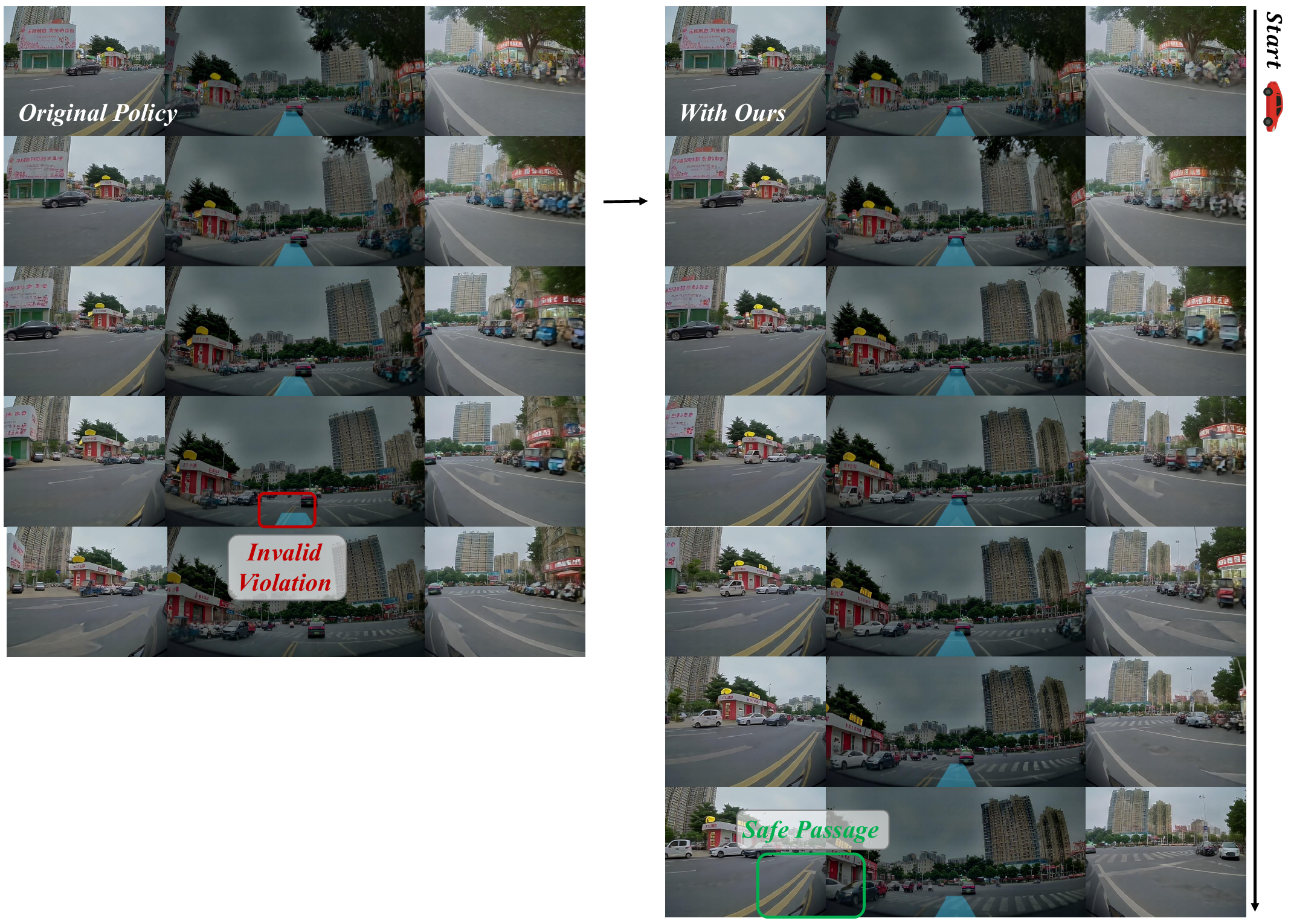} 
    \caption{Qualitative results of \ournet~avoiding lane violations on the in-house dataset.}
    \label{fig:supp_vis2}
\end{figure}

\begin{figure}[h]
    \centering
    \includegraphics[width=\linewidth]{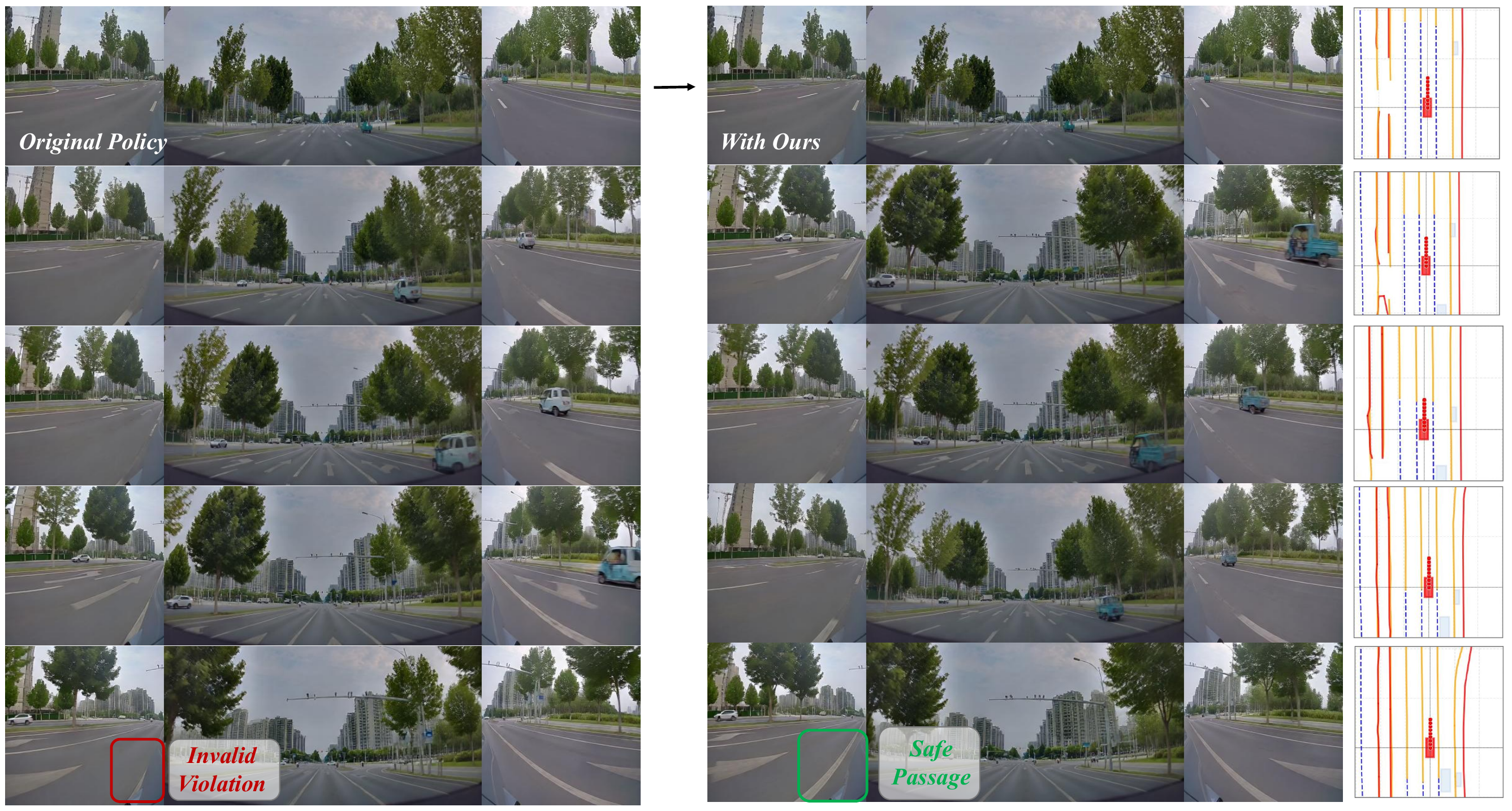} 
    \caption{Qualitative results of \ournet~avoiding lane violations on the in-house dataset.}
    \label{fig:supp_vis5}
\end{figure}

\begin{figure}[!h]
    \centering
    \includegraphics[width=\linewidth]{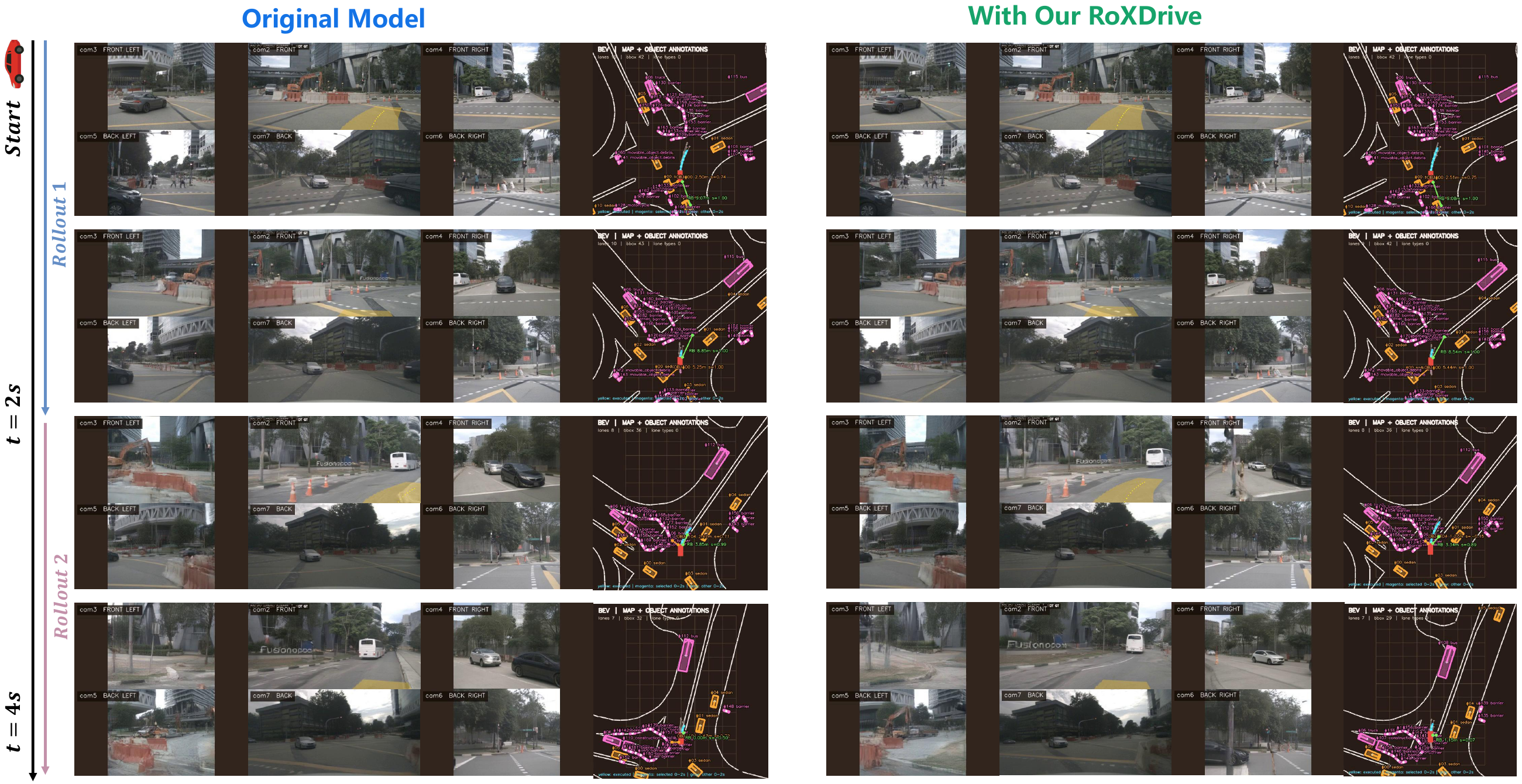}
    \caption{Additional qualitative results of \ournet~mitigating the causal confusion on nuScenes.
    }
    \label{fig:supp_vis6}
\end{figure}

\begin{figure}[!h]
    \centering
    \includegraphics[width=\linewidth]{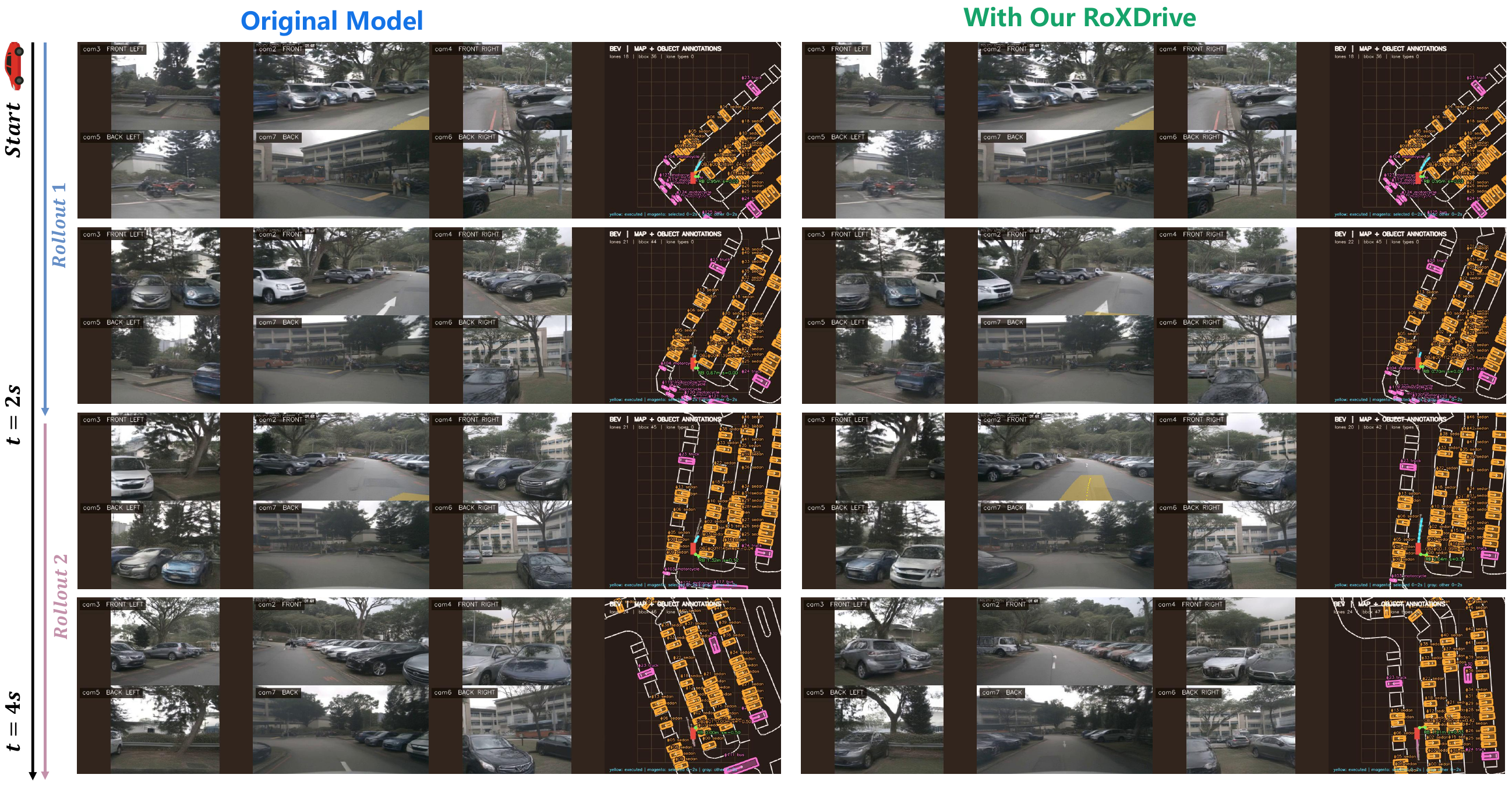}
    \caption{Additional qualitative results of \ournet~mitigating the causal confusion on nuScenes.
    }
    \label{fig:supp_vis7}
\end{figure}

 In this part, we provide additional qualitative results showing how \ournet~improves the model by avoiding collisions and lane-boundary violations, as shown in Figures~\ref{fig:supp_vis1}, \ref{fig:supp_vis3}, \ref{fig:supp_vis2}, \ref{fig:supp_vis5}, \ref{fig:supp_vis6}, \ref{fig:supp_vis7}, \ref{fig:supp_vis8} and \ref{fig:supp_vis9}.
We strongly recommend that readers watch the video demos in the supplementary materials.

\section{Limitations and future directions}
\label{supp_lim}
\noindent
\textbf{Reliability of world models and real-world deployment.}
Although \ournet~enables action-vision faithfulness evaluation, ego-action faithfulness alone may not always guarantee the correctness of the entire simulated environment.
Thanks to the strong generalization capability of X-World, we achieve accurate scene-controlled video generation on nuScenes, even with fine-tuning on only hundreds of 20 s training scenes supplemented with model-predicted annotations (\ie label noise).
Currently, physically consistent scene evolution and realistic responses of surrounding agents to novel ego actions remain important challenges, particularly beyond the logged data distribution.
It would be better if future work extended simulation assessment beyond ego-motion consistency to scene geometry and multi-agent interactions, and verified the policy improvements under real-world driving conditions.

\noindent
\textbf{Scalability of closed-loop exploration.}
High-fidelity multi-camera video generation remains computationally
expensive, limiting both interaction horizons and exploration breadth.
This creates a practical trade-off between simulation fidelity
and the interaction scale, while rare events and long-horizon driving decisions usually need to be further explored.
Although \ournet~improves learning from reliable rollouts, this underlying bottleneck needs more consideration and discussion.

\clearpage

\begin{figure}[!h]
    \centering
    \includegraphics[width=\linewidth]{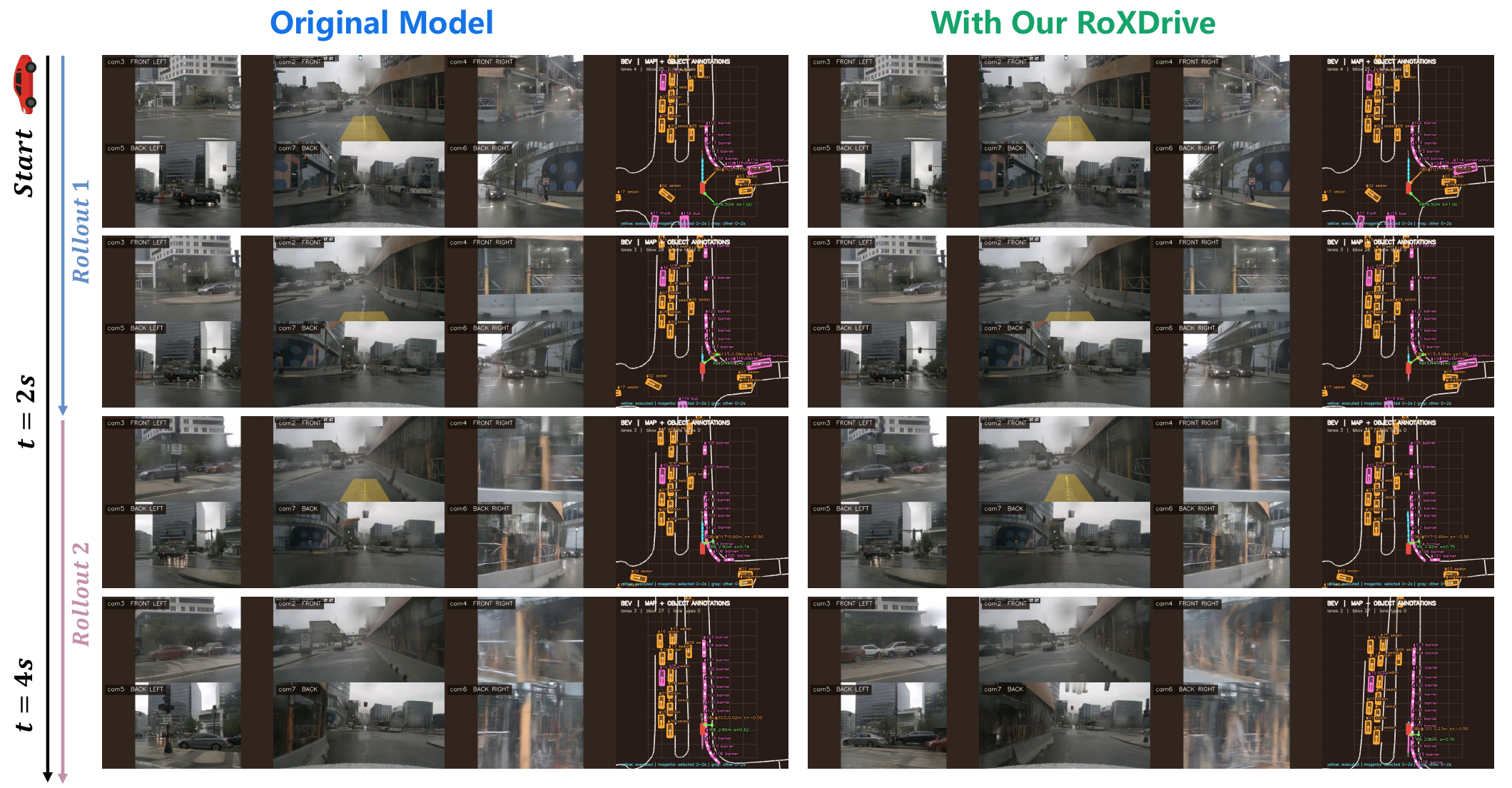}
    \caption{Additional qualitative results of \ournet~mitigating the causal confusion on nuScenes.
    }
    \label{fig:supp_vis8}
\end{figure}

\begin{figure}[!h]
    \centering
    \includegraphics[width=\linewidth]{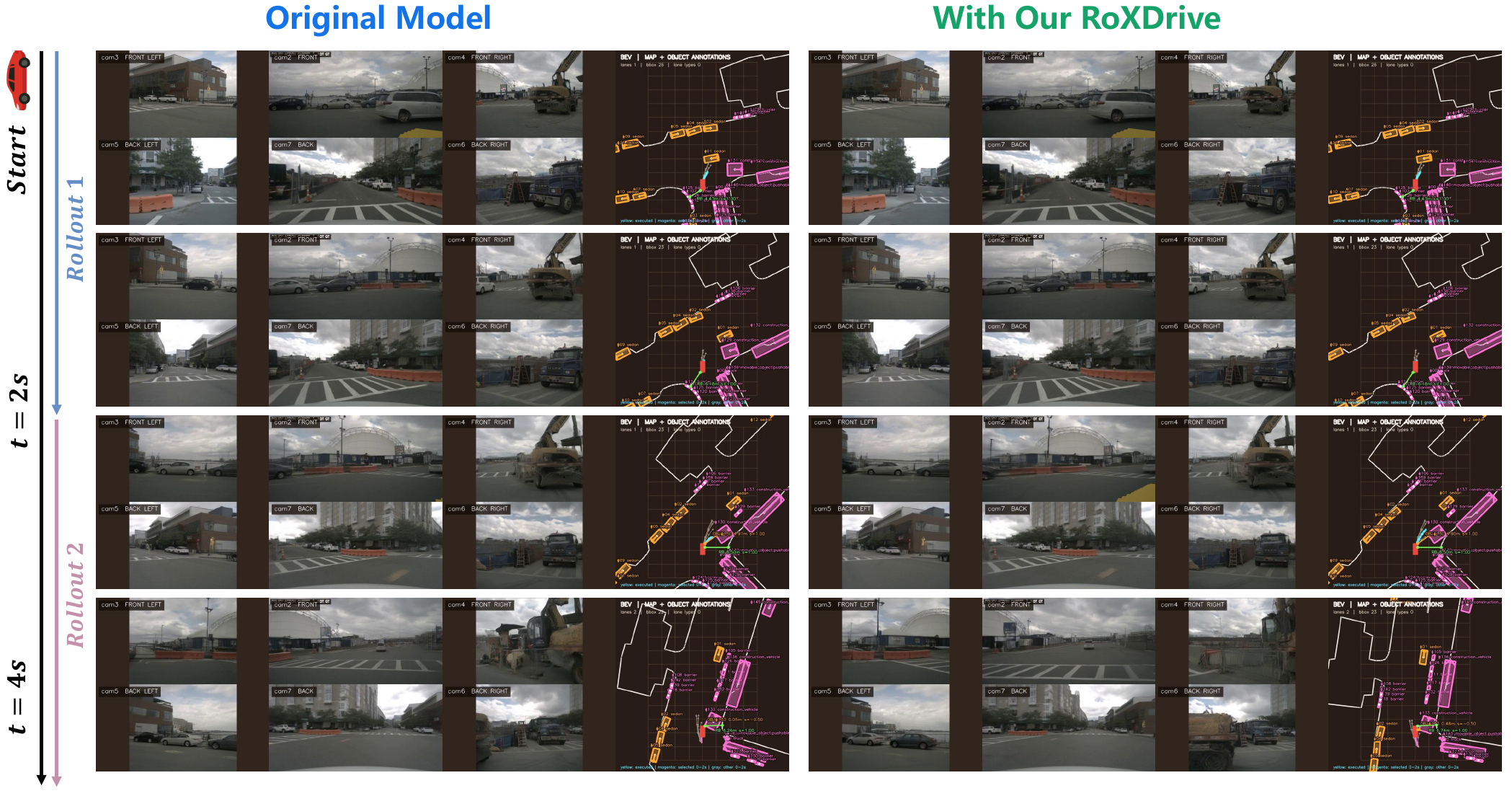}
    \caption{Additional qualitative results of \ournet~mitigating the causal confusion on nuScenes.
    }
    \label{fig:supp_vis9}
\end{figure}

\end{document}